\documentclass[accepted]{uai2022} %

\usepackage[british]{babel}

\usepackage{natbib} %
\usepackage{mathtools} %
\usepackage{booktabs} %
\usepackage{tikz} %
\usetikzlibrary{bayesnet}

\usepackage{subcaption}

\usepackage{xspace}
\usepackage{xcolor}
\usepackage{amsfonts}
\usepackage{physics,amsmath,amssymb}
\usepackage{array}
\usepackage{listings}

\DeclareMathOperator{\KL}{\ensuremath{KL}}

\newcommand{\ie}{\textit{i.e.}\xspace}
\newcommand{\eg}{\textit{e.g.}\xspace}

\newcommand{\enumone}{\textit{(i)}\xspace}
\newcommand{\enumtwo}{\textit{(ii)}\xspace}
\newcommand{\enumthree}{\textit{(iii)}\xspace}

\newcommand{\rsec}[1]{Section~\ref{sec:#1}}

\newcommand{\infovae}{\textsc{InfoVae} }
\newcommand{\betavae}{\textsc{BetaVae} }
\newcommand{\fbvae}{\textsc{FbVae} }

\usepackage{multirow}

\title{Statistical Model Criticism of Variational Auto-Encoders}

\author[1]{\href{mailto:<cbarkhof@gmail.com>?Subject=Statistical Model Criticism of Variational Auto-Encoders}{Claartje Barkhof}{}}
\author[1]{\href{mailto:<w.aziz@uva.nl>?Statistical Model Criticism of Variational Auto-Encoders}{Wilker Aziz}{}}
\affil[1]{%
    Institute for Logic, Language and Computation\\
    University of Amsterdam
}

\begin{document}
\maketitle

\begin{abstract}
We propose a framework for the statistical evaluation of variational auto-encoders (VAEs) and test two instances of this framework in the context of modelling images of handwritten digits and a corpus of English text.
Our take on evaluation is based on the idea of statistical model criticism, popular in Bayesian data analysis, whereby a statistical model is evaluated in terms of its ability to reproduce statistics of an unknown data generating process from which we can obtain samples.
A VAE learns not one, but two joint distributions over a shared sample space,
each exploiting a choice of factorisation that makes sampling tractable in one of two directions (latent-to-data, data-to-latent).  
We evaluate samples from these distributions, assessing their (marginal) fit to the observed data and our choice of prior, and we also evaluate samples through a pipeline that connects the two distributions starting from a data sample, assessing whether together they exploit and reveal latent factors of variation that are useful to a practitioner. 
We show that this methodology offers possibilities for model selection qualitatively beyond intrinsic evaluation metrics and at a finer granularity than commonly used statistics can offer.

\end{abstract}

\section{Introduction}\label{sec:intro}

Recent developments in 
 deep learning and approximate probabilistic inference \citep{mnih2014neural,TitsiasL14,Kingma+2014:VAE} 
have enabled the tractable estimation of flexible probabilistic models over complex sample spaces (\eg, images \citep{kingma2016improved}, natural language \citep{bowman2015generating}, molecules \citep{simonovsky2018graphvae}). 
Deep or shallow, statistical models are seldom short of inadequacies \citep{theis2015note}. Before a model can be assumed suitable to purpose, it is crucial that we criticise it along dimensions of relevance to the practitioner. 
Effective strategies for statistical model criticism \citep{box1980sampling} require a combination of statistics and knowledge of the application domain, but  methodology to assist in their design and automation does exist \citep{LloydEtAl2015}.  
This paper argues for the importance of statistical criticism in the context of deep probabilistic latent variable models (LVMs), in particular, one prominent class of deep LVMs---the variational auto-encoders \citep[VAEs;][]{Kingma+2014:VAE}---introducing  a methodology based on Bayesian estimation \citep{kruschke2013bayesian}. 

\citet{LloydEtAl2015} formulate criticism of a deep probabilistic model as a pipeline: choose a statistic, compute it for a data sample, use the probabilistic model as a null hypothesis and estimate a p-value in an attempt to reject the model.
One of the main innovations in their design is to use the model itself as a null hypothesis, an idea with roots in Bayesian model checking \citep{gelman1996posterior}.
Their other innovation concerns the first step of this pipeline: instead of hand-picking a statistic, they chose a kernel and let maximum mean discrepancy \citep[MMD;][]{gretton2012kernel} find a statistic under which the two samples are maximally discrepant. Whether this \emph{witness function} is even relevant will depend on the practitioner's ability to choose a good kernel. %
Altogether their strategy constrains model criticism to a  binary decision (\ie, is this model good under MMD or not?). Moreover, with a tool to find discrepancies in a rather flexible space of statistics, null falsification will eventually reject most models, for most models are imperfect along some view of the data. For criticism, we need tools that help uncover trade-offs of different models, rather than reject them as unable to account for the data's full complexity. Besides, we need tools that allow the analyst to control the degree of scrutiny that guides model comparison.

In this work, we derive a rich but low-dimensional statistic from the posterior predictive distribution of a latent structure model (\eg, a hierarchical Bayesian model such as a Bayesian mixture model or latent Dirichlet allocation \citep{blei2003lda}). This model is chosen by the practitioner depending on the specific view of the data they expect (or need) their VAEs to capture.
We also turn away from binary decisions and null hypothesis testing, instead comparing the statistics of different groups under a Bayesian mixed-membership model that allows for the comparison of multiple groups (\ie, a control group and as many model groups as we have competing VAEs) and can answer complex queries regarding posterior discrepancy amongst groups. We motivate different dimensions of criticism relevant to the evaluation of VAEs and carefully design control groups that allow for comparing models along those dimensions.
Finally, we demonstrate the application of this methodology in modelling the MNIST dataset \citep{deng2012mnist} and the sentences in the English Penn Treebank \citep{marcus1993ptb}.

\section{Background}\label{sec:background}

Estimation of a generative model involves associating a distribution $p_X$ with a random variable $X$ that captures an unknown data generating process $q_X$ for which we have observations $\mathcal D_X = \{x^{(n)}\}_{n=1}^N$.\footnote{\textbf{Notation:} we use capital letters (\eg, $X$, $Z$) for random variables and lowercase letters (\eg, $x$, $z$) for their assignments. For a random variable $X$, we use $\Omega_X$ to denote its domain and $p_X$ to denote its distribution (with some abuse of notation, we also use $p_X$ to denote the probability density function, but this should be unambiguous in context). Sometimes we need to refer to two distributions for the same random variable, each characterised by a different density function, in those cases we will use $p_X$ and $q_X$.} 
Commonly, we select from a parametric family the $p_X$ whose likelihood given observations is maximum. %
Statistical criticism of $p_X$ then involves determining whether \enumone $p_X$ approximates statistical properties of future data. 
In an LVM, $p_X$ is the marginal of a joint distribution $p_{XZ}$, where $Z$ is a latent variable. LVMs have various applications: a marginal  may be more expressive than simple parametric families available, %
which can lead to better fit of the data;
latent variables partition the probability space inducing clusters and/or other forms of potentially interpretable structure.
In a directed LVM, we choose a prior distribution $p_Z$ for the latent variable, which establishes its statistical and geometrical properties, and a conditional model $p_{X|Z=z}$ of the observed variable---the \emph{observational model}. %
In addition to \enumone, criticising an LVM involves determining whether \enumtwo the model reveals unobserved factors of variation (\ie, hidden structure) that are useful to a practitioner.  
Whereas \enumone concerns the fit of $p_X$ to the unknown data generating process, %
\enumtwo concerns the perceived usefulness of inferences based on the generative model's posterior distribution $p_{Z|X=x}$. Crucially, \enumtwo hinges on the observational model's ability to exploit the latent space to explain unobserved factors of variation of data samples.
Posterior inference for $p_{XZ}$ is generally intractable, hence modern LVMs use auxiliary components for approximate inference, these introduce additional nuances to model criticism.

VAEs \citep{Kingma+2014:VAE,RezendeEtAl14VAE} are directed LVMs that employ neural networks (NNs) to parameterise the observational model $p_{X|Z=z}$. This flexible parameterisation precludes marginal inference and thus standard gradient-based parameter estimation. VAEs resort to variational inference \citep[VI;][]{Jordan+1999:VI}, and amortised VI \citep{Kingma+2014:VAE} in particular, to circumvent intractable posterior inference and enable tractable parameter estimation. %
Amortised VI involves choosing a joint distribution $q_{ZX}$  determined by chaining an unknown data generating process $q_X$ %
and a conditional model $q_{Z|X=x}$ of the latent variable---the \emph{inference model}. Samples from the former are available through a dataset ($\mathcal D_X$), whereas the latter is a member of a tractable family parameterised by an NN and optimised along with the generative model $p_{XZ}$ to approximate that model's posterior distribution via minimisation of $\mathbb E_{x \sim \mathcal D_X}[\KL(q_{Z|X=x}||p_{Z|X=x})]$.
As a function of the inference model, the Kullback-Leibler divergence $\KL(q_{Z|X=x}||p_{Z|X=x})$ is independent of the  marginal density $p_X(x)$, thus yielding a tractable objective for the estimation of $q_{Z|X=x}$. As a function of the observational model, this leads to a lowerbound $\mathbb E_{z\sim q_{Z|X=x}}[\log p_{X|Z=z}(x)] - \KL(q_{Z|X=x}||p_Z)$ on the model log-likelihood (the evidence lowerbound, ELBO), which we can use as a proxy for the estimation of $p_{X|Z=z}$.
The inference model is a tool for tractable approximate inference, and in a VAE, it also enables estimation of $p_{XZ}$. %
Thus, the qualitative properties of $p_{Z|X=x}$ is affected by trade-offs we make in the specification of $q_{Z|X=x}$ (\eg, factorisation assumptions), properties of the objective of optimisation (\eg, KL divergence is asymmetric and, in the direction we use, it is lenient to underestimation of posterior variance), as well as our ability to optimise non-convex objective functions (\eg, practical gradient-based optimisers offer at most local convergence).
Even though $p_{XZ}$ and $q_{ZX}$ are meant to give two views of one probability space, these practical limitations lead the two spaces to diverge from one another. 
A necessary condition for their consistency is that their marginals match \citep{Song2020Understanding}. %
Thus, criticising a VAE further involves \enumthree assessing the extent to which %
chaining its trainable components recovers $q_X$ and $p_Z$ in expectation.

\begin{figure*}[t]
    \centering
    \includegraphics[width=\textwidth]{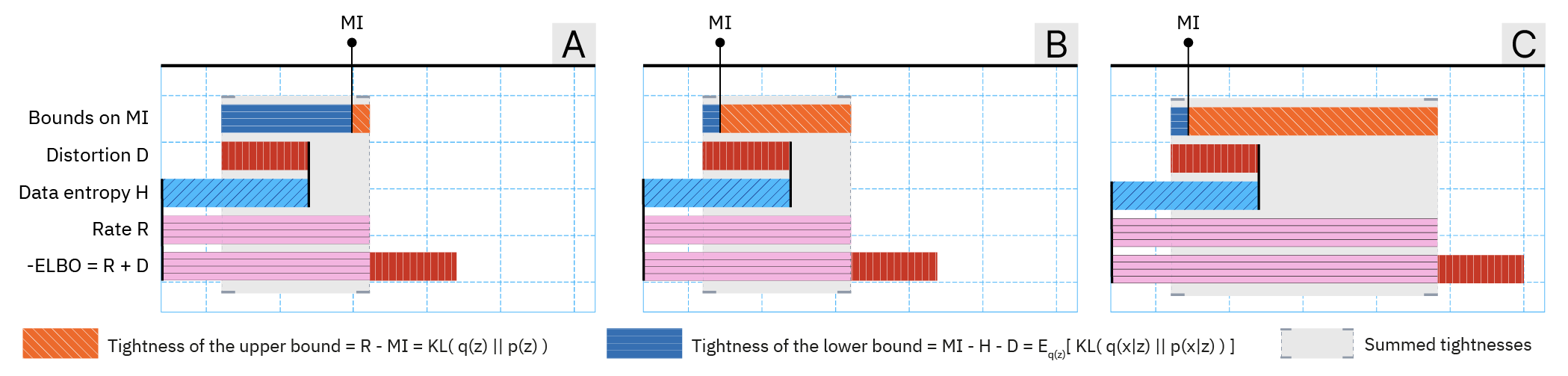}
    \caption{The diagrams illustrate three scenarios with different trade-offs in information theoretic quantities. Moving between scenario A and B keeps the ELBO, rate \textit{and} distortion fixed, while the level of MI varies within the range defined by its variational bounds (denoted with the grey area). This variability allows for one scenario (B) to have a higher level of marginal KL relative to the other scenario (A). This illustrates that marginal KL is an axis of variation not accounted for in the information theoretic RD-view on VAEs. Moving from scenario B to C does \textit{not} keep the RD-view (and thus ELBO) fixed and shows the potential hazardous situation where rate is elevated merely at the cost of marginal KL, without effectively diminishing distortion nor increasing MI. This scenario may seem predictably worse (after all, ELBO is worse), but it shows a caveat of techniques that focus on targeting higher rate solutions as a proxy to higher MI.}
    \label{fig:marginal-kl-diagram}
\end{figure*}

\section{Intrinsic Criticism of VAEs}\label{sec:intrinsic}

The most common intrinsic evaluation metric for density estimation is the log-likelihood (LL) of the model given a heldout sample of the data. The LL of a VAE is intractable, hence it is common to resort to Monte Carlo (MC) estimates through importance sampling \citep[IS;][]{robert2004monte}. 
While importance weighted LL \citep[IW LL;][]{burda2015importance} gives insight into the extent to which the marginal $p_X$ matches the implicit distribution $q_X$ on average  (criterion \enumone), it falls short in providing intuition for trade-offs made in the model regarding the usage of $Z$ that we practically care about (criteria \enumtwo \& \enumthree).

One example of a failure mode regarding criterion \enumtwo that need not be detected by IW LL is that of a model $p_{XZ}$ in which the latent variable $Z$ and observed variable $X$ are independent. This has been reported to occur in cases where the observational model is flexible enough to model $X$ without making use of $Z$, which establishes the consequence of a true posterior $p_{Z|X=x}$ that is independent of the data and is said to have \textit{collapsed to the prior} \citep{chen2016variational, bowman2015generating}. Due to this independence, VI finds a trivial optimum where $q_{Z|X=x}$ reduces to $p_Z$, perfectly recovering  the collapsed $p_{Z|X=x}$ for every $x \in \Omega_X$.  %
Motivated by this observation, \citet{alemi2018fixing} develop a view over the two axes the expected ELBO naturally decomposes into: the rate-distortion (RD) plane.\footnote{The expected ELBO can be rewritten as $-D - R$, where $\mathbb E_{x \sim q_X}[\KL(q_{Z|X=x}||p_Z)]$ is the rate $R$ and the distortion $D$ is $\mathbb E_{x \sim q_{X}, z \sim q_{Z|X=x}}[\log p_{X|Z=z}(x)]$.} They demonstrate that these two quantities parametrically determine the unique variational bounds between which the mutual information (MI) between the observed variable $X$ and latent variable $Z$ under $q_{ZX}$ is guaranteed to exist for a given parametric family. This leads to the insight that for any given ELBO level, the RD ratio may vary and so may the level of MI (quantitatively associated with criterion \enumtwo). 

While the RD decomposition and its relation to the bounds on MI leads to an expanded intrinsic evaluation with regards to the goals outlined in \rsec{background}, a closer look at these bounds reveals that there is a degree of freedom the decomposition through the lens of $q_{ZX}$ does not account for, and which is directly relevant to criterion \enumthree: $q_Z$ matching $p_Z$. This can be quantified as the KL from the prior $p_Z$ to the marginal $q_Z$ \citep[a.k.a. \textit{marginal KL};][]{hoffman2016elbo} and can be interpreted as the tightness of the upperbound on MI, the rate, to the true MI. And, importantly, it may vary relative to the tightness of the lowerbound at fixed ELBO \textit{and} fixed rate-distortion ratio (see Figure \ref{fig:marginal-kl-diagram}A \& B).
Furthermore, let us move from the theoretical situation where we reason in terms of fixed ELBO levels to a more realistic scenario where the optimal ELBO level at different RD ratios for a given parametric family is more likely to be defined by a bent RD-curve. In this scenario, the marginal KL may vary \textit{independently} of the tightness of the lowerbound on MI. This is practically relevant as there are numerous optimisation techniques that directly or indirectly aim at solutions that lay on segments of or form points on the RD contour defined for a parametric family \citep[e.g. ][]{kingma2016improved, pelsmaeker2019effective, chen2016variational, alemi2018fixing}. This thus potentially comes at the cost of compromised consistency and translates to higher marginal KL (Figure \ref{fig:marginal-kl-diagram}C).
Considering a more complete view of RD together with marginal KL or MI can still lead to difficulties: it is hard to obtain good estimates of those quantities  due to their dependence on the marginal $q_Z$, which is expensive to estimate and empirically bounded by $\log N$ \citep{Song2020Understanding}. %
On the other hand, even if we had perfect estimates at our disposal, reasoning on trade-offs between these quantities or proxies such as MMD is hard. And lastly--and possibly most importantly--none of these intrinsic metrics are explicitly designed to capture characteristics a practitioner may practically care about.

\section{Statistical criticism for VAEs}\label{sec:approach}

Our approach to comparing multiple VAEs is based on statistical criticism through Bayesian estimation \citep{kruschke2013bayesian,benavoli2017time}. We observe measurements from different groups of interest (\ie, relevant statistics of data  and/or model samples), some of which serve as control groups to help establish degrees of reasonable variation. We posit a hierarchical model of all grouped measurements and infer a posterior distribution over its latent parameters. We then use the posterior distribution to visualise and/or quantify degrees of discrepancy between any model group and a control group, and use such summaries to, for example, order model groups in terms of resemblance to the control group.
We have three criteria along which we criticise VAEs, namely,
\enumone `does $p_{X}$ fit the data?', \enumtwo `is $p_{Z|X=x}$ practically useful', \enumthree `do $p_{XZ}$ and $q_{ZX}$ offer two views of the same probability space?'. Next, we describe our approach to comparing VAEs along these three dimensions.

\paragraph{Strategy.} To criticise VAEs along criterion \enumone we compare how a real-valued statistic $T(X)$ distributes as a function of data samples $x \sim q_X$ or model samples obtained by chaining the prior and the observational model ($\hat x \sim p_{X|Z=z}$ with $z \sim p_Z$). If $T(X)$ and $T(\hat X)$ distribute similarly, then the VAEs correlate the variates in $X$ at least in the way $T(\cdot)$ does.
For criterion \enumtwo, we design a conditional statistic $T(X'|X)$ which has a structured view of $(X,X')$ and compare how it distributes as a function of data samples $x' \sim q_X$ or model samples obtained by chaining the inference model and the observational model ($\tilde x \sim p_{X|Z=z}$ with $z \sim q_{Z|X=x}$) given a seed data sample $x \sim q_X$. If $T(X'|X)$ and $T(\tilde X|X)$ distribute similarly, the VAEs use latent space to correlate $X$ and $\tilde X$ \emph{at least} to the extent that $T(\cdot|\cdot)$ does.
For criterion \enumthree, we design two diagnostics. First, we go back to the unconditional statistic $T(X)$ and compare its distribution to that of $T(\tilde X)$. Conditional model samples $\tilde x$ are model-based replications of seed samples, and, in expectation under $q_X$, they should reproduce patterns of the marginal $p_X$, no matter what the latent space is used for (or if it is used at all).
Second, we turn to latent space, which, unlike $\Omega_X$, is a (relatively) low-dimensional space, hence we compare prior samples ($z \sim p_Z$) to marginal samples ($\hat z \sim q_{Z|X=x}$ with $x \sim q_X$) directly. %

\paragraph{Data samples.} We assume availability of two collections drawn from $q_X$, one we shall call \emph{training data} and denote $\mathcal D_X$, one we shall call \emph{heldout data} and denote $\mathcal H_X$. 
Training data are called as such for they overlap with the data used for parameter estimation of the VAEs themselves.
Both data sets are available for the comparison of VAEs, but, as VAEs are point-estimated using $\mathcal D_X$, heldout data help us assess their ability to generalise beyond training data.

\paragraph{Statistics.} 
We use a real-valued statistic, rather than performing the analysis in data space directly, because it abstracts away from the high dimensionality of the data and because it can be made sensitive to a specific structured view of the data, one the practitioner is interested in.
By the latter we mean, the practitioner is interested in models whose samples are indistinguishable from data samples at least through the lens of $T(\cdot)$ and/or whose latent spaces capture at least the structure that $T(\cdot|\cdot)$ is sensitive to.
Concretely, we derive both statistics from the posterior distribution of a Bayesian latent structure model $\mathcal S$ whose modelling assumptions the practitioner controls. %
We condition on data samples available for the analysis (\eg, $\mathcal D_X$) and expose their structure through the lens of the latent parameters of $\mathcal S$.
Finally, for a future outcome $x_*$, $T(x_*)$ is the logarithm of the posterior predictive density (lppd) under $\mathcal S$: $T(x_*) = $
\begin{equation}\label{eq:lppd}
    \log \int  \sum_{c_*} p_{\text{samp}}(x_*|\phi, c_*)p_{\text{post}}(\phi, c_*|\mathcal D_X) \dd\phi ~,
\end{equation}
where $p_{\text{post}}$ is the posterior of the analysis model, $p_{\text{samp}}$ is that model's sampling distribution, $\phi$ is a global latent parameter and $c$ is a local latent variable (typically discrete). 
This statistic can be thought of as a measure of discrepancy between a future outcome and the model $\mathcal S$---that is, the explicit modelling assumptions it makes and the data it conditions on. %
In practice we use a sampled estimate of the marginal probability.
From the same model, we can derive the conditional statistic $T(\tilde x_*|x_*)$ of a replication $\tilde x_*$ of a seed sample $x_* \sim \mathcal H_X$ by assessing lppd under $\mathcal S$'s posterior distribution updated to also condition on the seed $x_*$: $T(\tilde x_*|x_*) = $
\begin{equation}\label{eq:rlppd}
    \log \int  \sum_{c_*} p_{\text{samp}}(\tilde x_*|\phi, c_*)p_{\text{post}}(\phi, c_*|\mathcal D_X, x_*) \dd\phi ~.
\end{equation}

\paragraph{Bayesian estimation and model comparison.} We are not interested in the magnitude of lppd as such, rather we aim at quantifying discrepancy in how this score distributes for each model group relative to a control group. 
As we shall see in \rsec{experiments}, the statistics from different groups distribute in rather complex ways, thus, rather than fitting a simple parametric family (\eg, a Student's $t$, as done in \citep{kruschke2013bayesian}), we use a Bayesian mixed-membership model \citep{blei2014build,airoldi2015handbook} to jointly infer flexible density functions for all groups being analysed. 

Let $\mathbf y \in \mathbb R_{>0}^I$ denote all measurements for the $G$ groups in the analysis, each measurement being the negative of an outcome's lppd under $\mathcal S$. A grouped mixed-membership model samples the parameters of $K$ components $(\mu_1, \sigma_1), \ldots, (\mu_K, \sigma_K)$ along with $G$ vectors of mixing coefficients $\boldsymbol\pi_1, \ldots, \boldsymbol\pi_G$, each $\boldsymbol\pi_g \in \Delta_{K-1}$, from a Dirichlet process (DP) prior (\ie, $K \to \infty$) with concentration $\alpha > 0$ and a base measure appropriate for the type of measurement. To model the negative of an outcome's lppd, we use a mixture of truncated Normal distributions $\mathcal N_{+}$ and our base measures are uniform over large subsets of the positive real line (details in appendix Section \ref{app:dp-models}). The likelihood function of the model is $\prod_{i=1}^I f(y_i|\boldsymbol\pi_{g_i},\mu,\sigma)$, where $g_i \in \{1, \ldots, G\}$ indicates the group to which $y_i$ belongs and $f(y|\boldsymbol\pi,\mu,\sigma) = \sum_{k=1}^K \pi_{k} \mathcal N_{+}(y|\mu_{k}, \sigma_{k}^2)$. 
To model latent variables directly, we use a mixture of low-rank multivariate Normal distributions and adjust the base measures accordingly (details in the appendix Section \ref{app:latent-structure-models}).

\paragraph{Statistical discrepancy}
Our goal is to quantify discrepancy in distribution between groups of measurements $\{y_i: g_i = c\}$ and $\{y_i: g_i = m\}$, a control group $c$ and a model group $m$, given measurements for all groups $\mathbf y$. 
For that we approximate the posterior KL divergence from the mixture $f_m(y_*) \triangleq f(y_*|\boldsymbol\pi_m, \mu, \sigma)$ to the mixture $f_c(y_*) \triangleq f(y_*|\boldsymbol\pi_c, \mu, \sigma)$: $\mathbb E_{}[\KL(f_c || f_m) | \mathbf y] \le$
\begin{equation}
    \mathbb E_{}[\KL(\mathrm{Categorical}(\boldsymbol\pi_c) || \mathrm{Categorical}(\boldsymbol\pi_m)) | \mathbf y] ~,
\end{equation}
where the KL between a mixture of shared components is upperbounded by the KL divergence between the distributions over component assignments \citep{hershey2007approximating}. 
We obtain an MCMC estimate of the expected value, for the DP we truncate a stick-breaking procedure at finite $K$. We make use of a No-U-Turn Sampler for improved efficiency \citep{hoffman2014nuts}. For the analysis of $Z$ we resort to SVI instead of MCMC.

\section{Experiments}\label{sec:experiments}

In this section we  demonstrate the use of the proposed statistical framework for evaluating a set of VAEs in the context of modelling a statically binarised version of MNIST \citep{deng2012mnist} and the English PTB \citep{marcus1993ptb}.

\subsection{VAEs}

\paragraph{Optimisation criteria.} We employ three different optimisation criteria in our experiments that by design relate to the issues highlighted Section \ref{sec:intrinsic}. 
\textbf{$\beta$-VAE} (\textsc{BetaVae}): \citet{higgins2016beta} propose a framework for controlling the capacity of the information bottleneck by adding a hyperparameter $\beta$ to the ELBO objective. \textbf{Info-VAE} (\textsc{InfoVae}): \citet{zhao2017infovae} propose this optimisation criterion to allow for explicitly controlling the balance between accurate posterior inference and maintaining substantial mutual information between the latent and observed variables. We implement this objective as a $\beta$-VAE objective with an additional MMD term, weighted with hyperparameters $\lambda_\text{rate}$ and $\lambda_{\text{MMD}}$ respectively. \textbf{Free bits VAE} (\textsc{FbVae}): to counteract the mutual information between the latent and observed variables to vanish and indirectly stimulate a non-collapsed posterior \citet{kingma2016improved} propose to alter the expected ELBO replacing $R$ by $\max(\lambda_\text{FB}, R)$ in an attempt to enforce a minimum rate.
For both the experiments on MNIST and PTB we select a range of values for $\beta$, $\lambda_\text{rate}$, $\lambda_{\text{MMD}}$ and $\lambda_\text{FB}$, which are listed together with an overview of the criteria in the supplementary material (Section \ref{app:hyperparameters}, Tables \ref{tab:objectives-hp} and \ref{tab:objectives}).

\paragraph{Architectures.} For all experiments we implement a standard VAE with a fully factorised Gaussian approximate posterior $q_{Z|X=x}$ and a standard Gaussian as prior $p_Z$. The dimensionality of the latent space used across all the MNIST experiments is 10 and for all the PTB experiments 32. We describe some important architectural details below and refer the reader to Section \ref{app:architectures} of the supplementary material for more details. {\bf MNIST:} We experiment with two types of observational model. The simple model is a fully factorised product of Bernoulli distributions with gated transposed convolution layers as its main building block (\ie, CNN.T decoder), following \citet{van2018sylvester}. %
The more complex observational model employs a conditional PixelCNN++ architecture \citep{salimans2017pixelcnn++} to achieve an autoregressive product of Bernoulli distributions, following the implementation of \cite{alemi2018fixing}. %
We keep the architecture of the inference model (the \textit{encoder}) fixed and again follow the implementation with gated convolutional layers of \citet{van2018sylvester}. {\bf PTB:} We experiment with an auto-regressive factorisation of the observational model, and follow \cite{li2020optimus} in altering a transformer architecture \citep[RoBERTa;][]{liu2019roberta} to an auto-regressive model that incorporates the latent via two mechanisms: the embedding mechanism and the attention mechanism. To decrease computational overhead we initialise the weights with a distilled checkpoint \citep{Sanh2019DistilBERTAD}. For the encoder we use an original version of the RoBERTa architecture initialised with the same distilled weights.

\subsection{Latent structure models}\label{subsec:bda_models}

Here we describe the latent structure models (LSMs) used to assign lppd to the model samples and control groups outlined in \rsec{approach}. 
Their designs and complexity may vary across application domains and be chosen according to the practitioners liking, but we encourage to iteratively build complexity as even simple models can have surprising discriminative power, as we shall see. The graphical representations of our LSMs are shown in Table \ref{tab:BDA_diagrams} in the supplementary material. %
It is worth noting that even though LSMs themselves are of generative nature, their ability to act as a generator in a realistic scenarios is simply too limited. But, their latent parameters may capture a data-driven notion of the latent structure at hand which a practitioner may want a more powerful  generator, such as a VAE, to also induce. Moreover, if samples from a VAE do not seem data-like under the LSM's (simple) factorisation, the VAE has surely failed at modelling the complex $q_X$.

\paragraph{MNIST digit identiy model.}
The most obvious latent structure present in the MNIST dataset is that of the digit identity. Assessing whether VAEs capture this structure helps detect severely failed optimisation. To this end we design a simple LSM that is a mixture over 10 components, each of which is defined as a joint over independent Bernoulli distributions with a shared Beta prior to model the pixel values. This can be thought of as a naive Bayes classifier, and, in fact, we supervise it by pre-annotating the MNIST training data with classes predicted using a k-nn classifier (with 96.6\% accuracy).

\paragraph{PTB sequence length model.}

A core property of written text is variability in sequence length. To test whether our VAEs capture this characteristic credibly, we design an LSM that models sentence length under a latent mixture of Poissons (latent components in Figure \ref{fig:ptb_data_model_length_distributions} of the supplement). %

\paragraph{PTB topic model.}

Subsequently, we define another LSM to evaluate VAEs fitted on PTB which focuses more on the content of the written text. To this end, we use latent Dirichlet allocation \cite[LDA;][]{blei2003lda} to uncover an underlying topic structure to the text we are modelling in the form of distributions over word count vectors. %

\subsection{Analysis}

\begin{figure}
    \centering
    \includegraphics[width=0.5\textwidth]{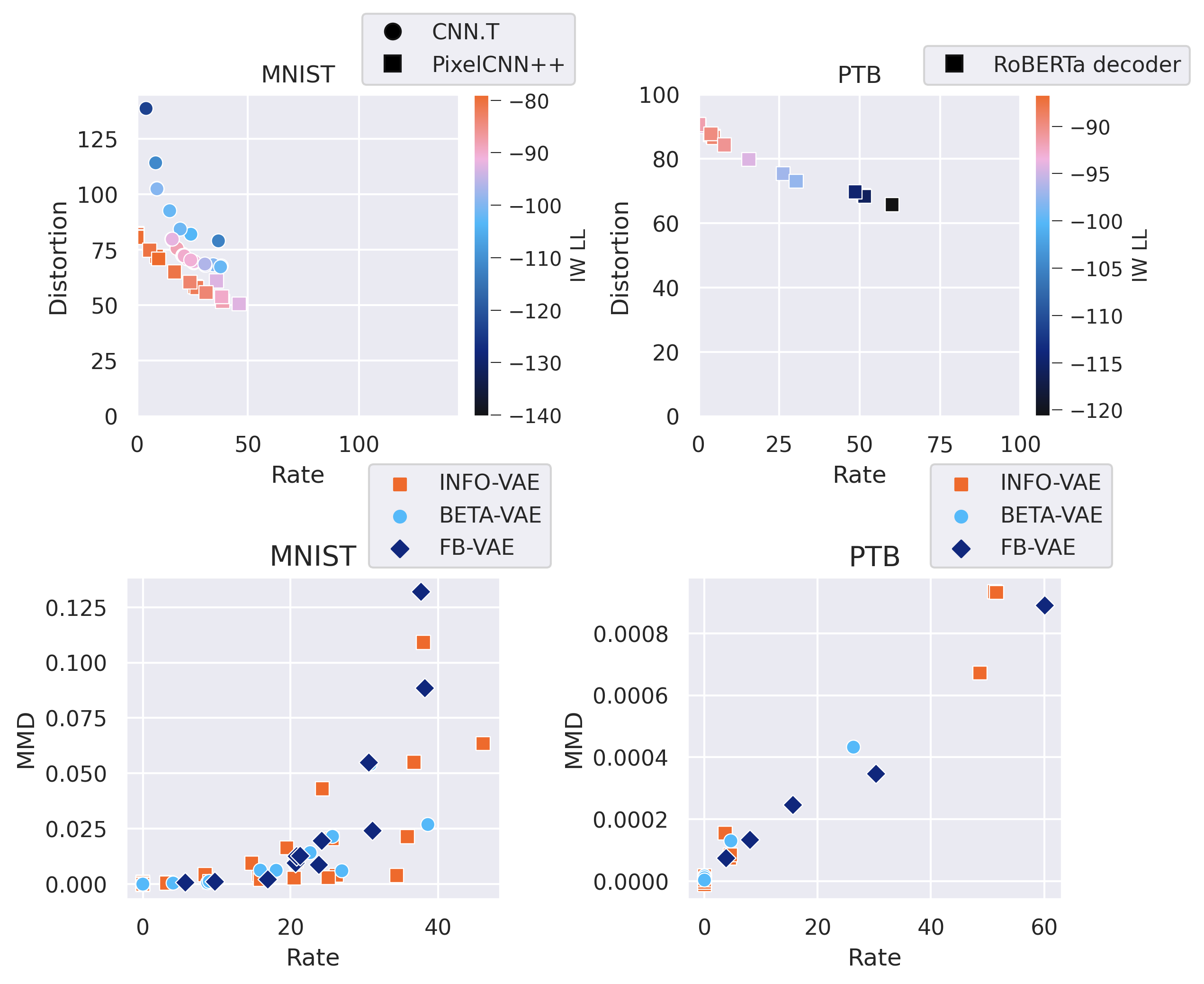}
    \caption{The intrinsic evaluation results of all experiments.}
    \label{fig:intrinsic-evaluation-plot}
\end{figure}

\paragraph{Intrinsic evaluation.}
Intrinsic evaluation metrics are summarised visually for all experiments in Figure \ref{fig:intrinsic-evaluation-plot}. Full tabular results can be found in the supplementary material (Section \ref{app:intrinsic-evaluation}). For the MNIST CNN.T experiments we see that the trade-off between R and D is defined by a bent curve: the first segment of the curve defines a region where D can be diminished with R efficiently (steep D decrease between experiments) and the second segment of the curve defines a region where the opposite is true. For the PixelCNN++ experiments and all the PTB experiments we observe more typical strong decoder behaviour where the conversion between R and D is an inefficient one for all levels of R that have been recorded in our experiments. In the lower two plots we can observe that for almost all experiments MMD increases with an increase in R. To which extent this happens differs per objective in the MNIST experiments, but for PTB it seems that all objectives act quite similarly in this regard. Additionally, we observe that the MMD scale is quite different for MNIST than for PTB, which makes it hard to tie practical judgements to the consequence of elevated MMD.

\begin{figure*}[t]
     \centering
     \begin{subfigure}{\textwidth}
         \centering
         \includegraphics[width=\textwidth]{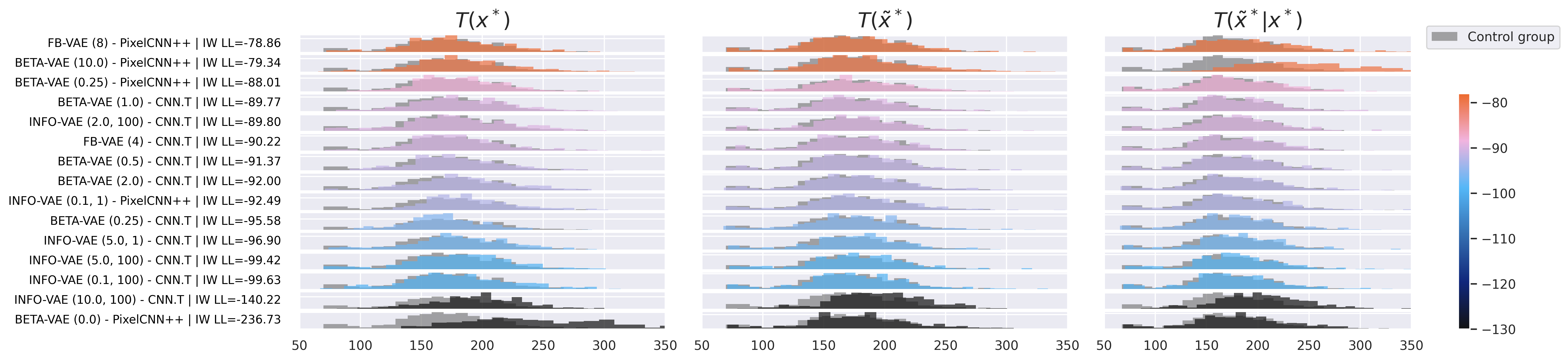}
         \caption{MNIST digit identity}
         \label{fig:all-surprisal-dists-sub-mnist}
     \end{subfigure}
     \begin{subfigure}{\textwidth}
         \centering
         \includegraphics[width=\textwidth]{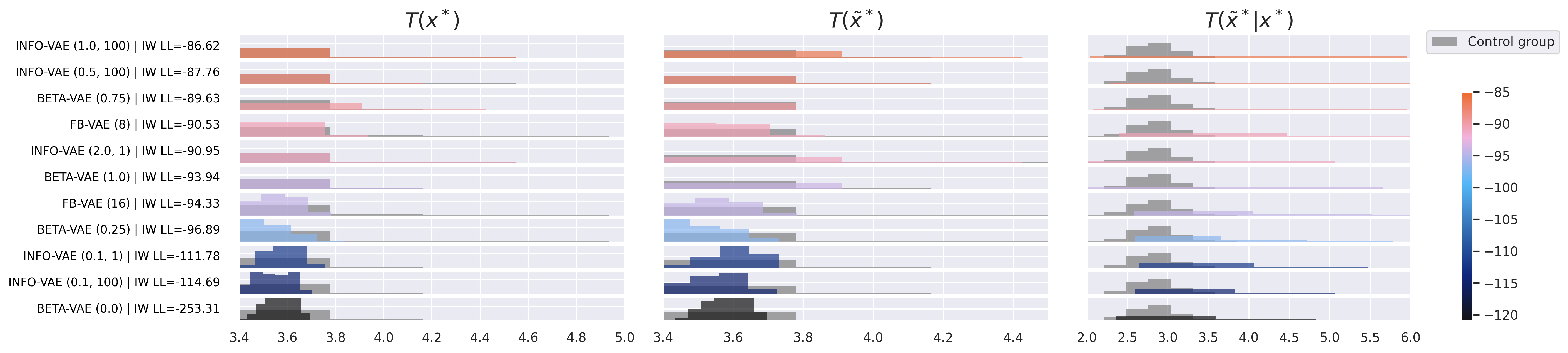}
         \caption{PTB sequence length}
         \label{fig:all-surprisal-dists-sub-ptb-seq-len}
     \end{subfigure}
     \begin{subfigure}{\textwidth}
         \centering
         \includegraphics[width=\textwidth]{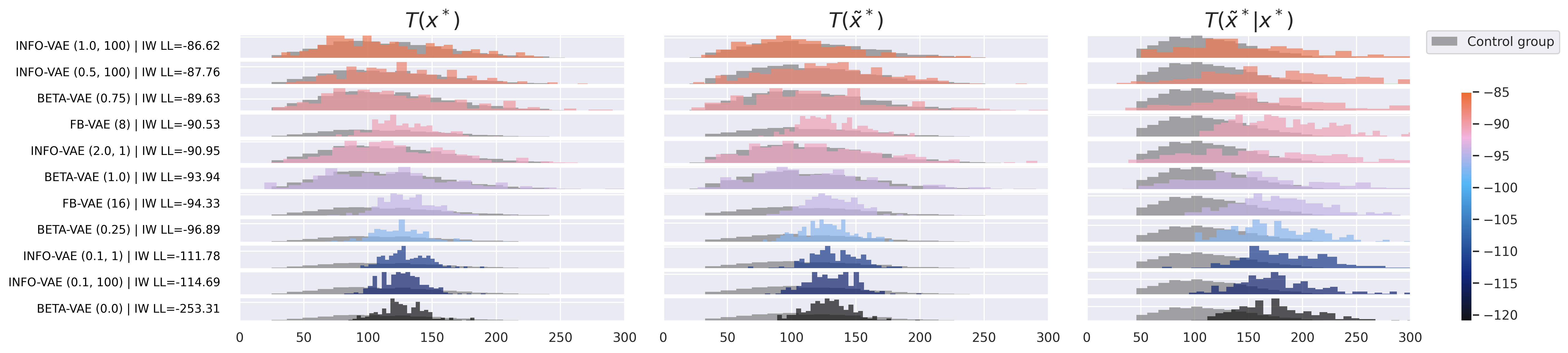}
         \caption{PTB topics}
         \label{fig:all-surprisal-dists-sub-ptb-topics}
     \end{subfigure}
        \caption{The three statistics $T(X_*)$, $T(\tilde X_*)$ and $T(\tilde X_*|X_*)$ assessed under the three LSMs visualised for a subset of the experiments together with the distributions of the control groups. The rows are ordered and coloured by IW LL. The experiments are labelled with the objectives according to the following format: \infovae ($\lambda_{\text{rate}}$, $\lambda_{\text{MMD}}$), \betavae ($\beta$) and \fbvae ($\lambda_{\text{FB}}$). For the MNIST experiments we additionally distinguish between decoder types used (CNN.T or PixelCNN++). We refer the reader for full experimental results to the supplementary material (Section \ref{app:latent-structure-models}).}
        \label{fig:all-surprisal-dists}
\end{figure*}

\paragraph{Negative lppd under latent structure model.}
To get an initial overview of the result of our analysis, we plot the histograms of the collected statistics for a subset of the experiments together with those belonging to the control group in Figure \ref{fig:all-surprisal-dists}. We sort and colour the rows of the plots according to IW LL. Plots with all experiments can be found in the supplementary material (Section \ref{app:latent-structure-models}). By inspecting the distributions we can make a few general observations. Primarily, it can be observed that the ordering according to IW LL does not generally correspond to the perceived divergences in histograms from the control group across statistics. On both ends of the IW LL spectrum we can find distributional discrepancies with respect to to the control group. Conversely, we can perceive differences in the distribution of statistics for models that are nearly identical in terms of IW LL. For the MNIST digit identity model (Figure \ref{fig:all-surprisal-dists-sub-mnist}), for example, we can see that the best models in terms of statistics distributions seem to reside around an IW LL of $-90$. In fact, we can even distinguish models in that range by inspecting the histograms closely. We can, for example, distinguish them in their ability to model the multi-modal nature of the control group with regards to $T(X_*)$: the \infovae (2.0, 100) MNIST experiment seems to capture the small chunk of probability mass at the lower end of the spectrum better than the experiment above and below it. Similarly, for the PTB topic model (Figure \ref{fig:all-surprisal-dists-sub-ptb-topics}) we can for instance visually appreciate differences between the \fbvae (8.0) experiment and its row-wise neighbours, while they have nearly identical average IW LL estimates.

\begin{figure*}[ht]
     \centering
     \begin{subfigure}{\textwidth}
         \centering
         \includegraphics[width=\textwidth]{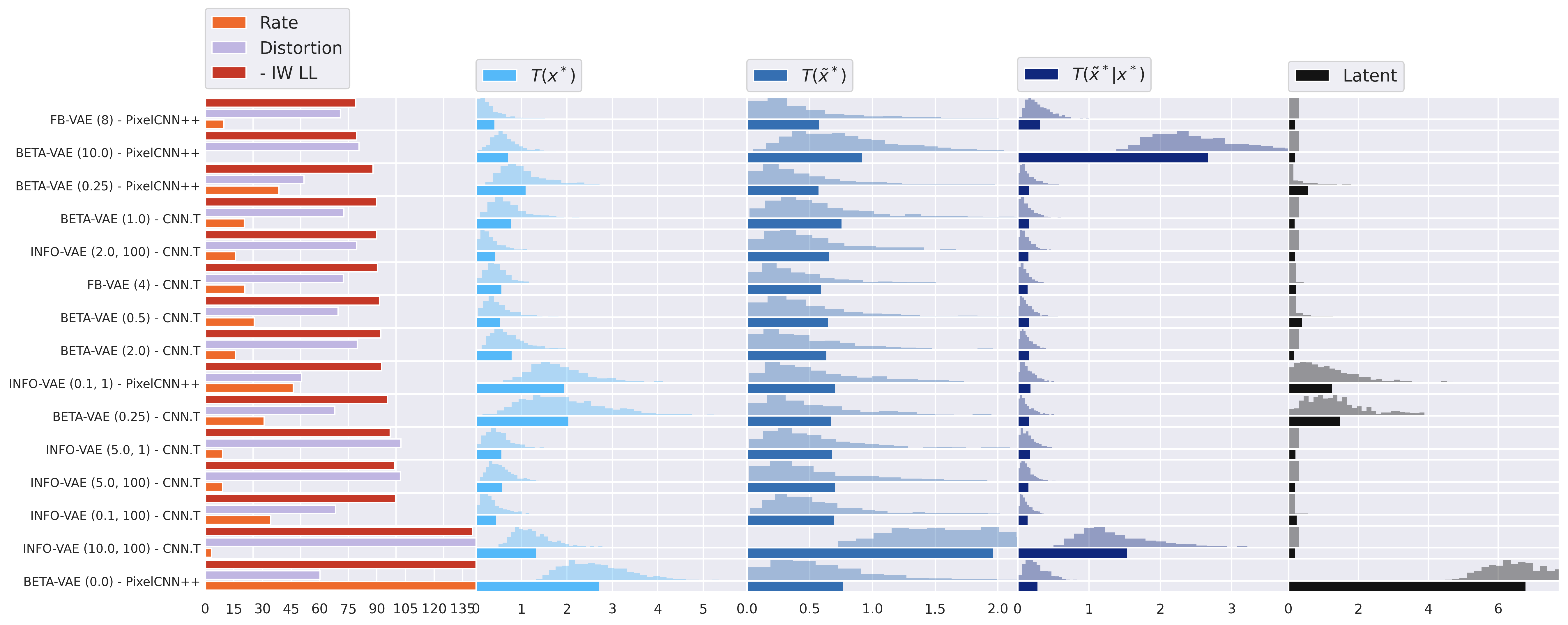}
         \caption{MNIST digit identity}
         \label{fig:all-kl-plots-sub-mnist}
     \end{subfigure}
     \begin{subfigure}{\textwidth}
         \centering
         \includegraphics[width=\textwidth]{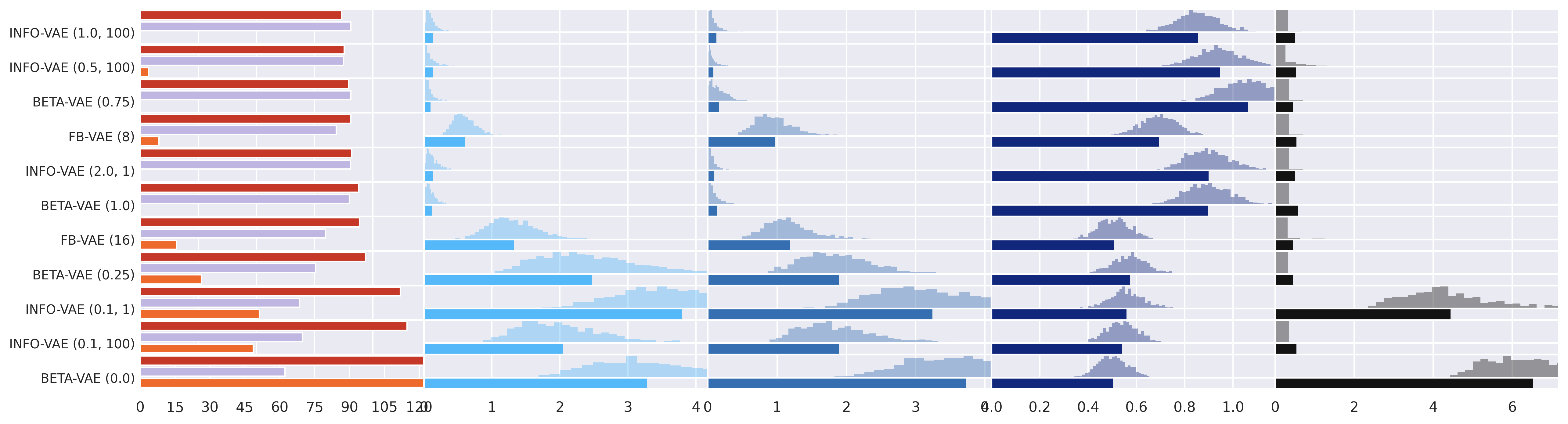}
         \caption{PTB sequence length}
         \label{fig:all-kl-plots-sub-ptb-seq-len}
     \end{subfigure}
     \begin{subfigure}{\textwidth}
         \centering
        \includegraphics[width=\textwidth]{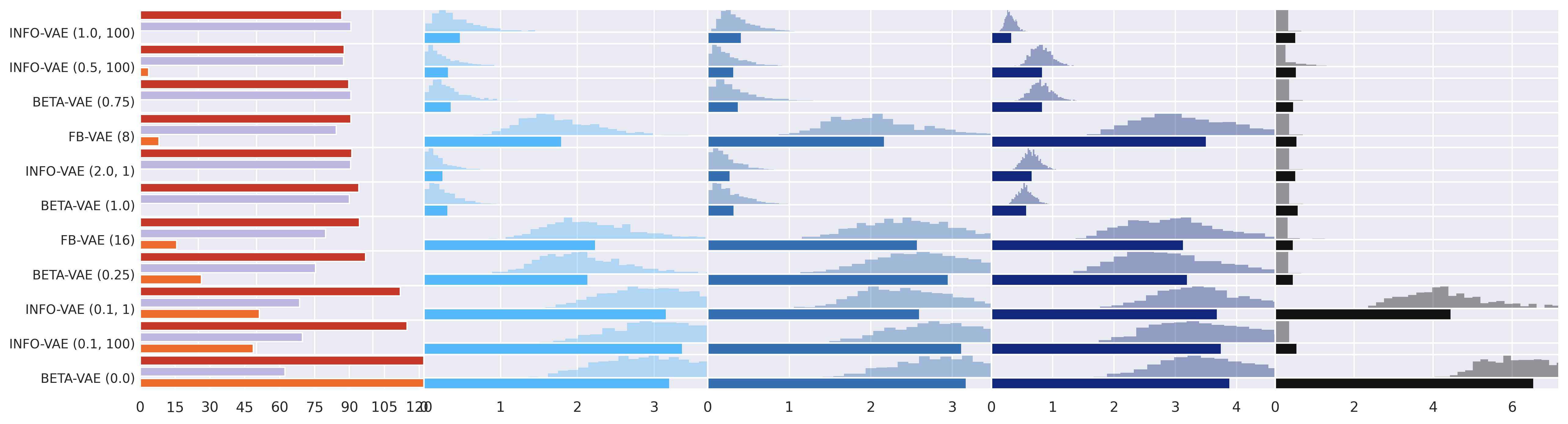}
         \caption{PTB topics}
         \label{fig:all-kl-plots-sub-ptb-topics}
     \end{subfigure}
        \caption{A summarising view of our analysis of a subset of the experiments. The leftmost column shows the intrinsic evaluation metrics for reference. The next three columns show estimated divergences from the control group under our analysis model. The rightmost column shows the estimated divergence from $q_Z$ to $p_Z$. Full experimental results can be found in the supplementary material (Section \ref{app:kl-plots}). The horizontal bars denote the average value of the sampled posterior divergences plotted as histograms. The experiments are labelled with the objectives according to the following format: \infovae ($\lambda_{\text{rate}}$, $\lambda_{\text{MMD}}$), \betavae ($\beta$) and \fbvae ($\lambda_{\text{FB}}$). For the MNIST experiments we additionally distinguish between decoder type used: CNN.T or PixelCNN++.}
        \label{fig:all-kl-plots}
\end{figure*}

\paragraph{Posterior KL upperbound.} Figure \ref{fig:all-kl-plots} is a summary of the analysis. It shows intrinsic evaluation metrics (1st column) alongside the estimated discrepancies relative to the control groups under the analysis models as described in Section \ref{sec:approach} for a subset of the experiments. Full results can be found in the supplementary material Section \ref{app:kl-plots}. The next three columns correspond to the discrepancies between distributions of the three lppd statistics in data space and the rightmost column shows the discrepancy between prior samples and samples from the inference model averaged under $\mathcal H_X$.
For the {\bf digit identity model}, we can observe that low R models have a high average divergence to the data group in terms of the conditional statistic $T(\tilde X_*|X_*)$. This aligns with the intuition that these models fail to capture information on the digit identity in the latent space. Additionally, we can observe that increased R often coincides with larger discrepancy along the $T(X_*)$ and $T(\tilde X_*)$ statistics. 
On {\bf PTB sequence length}, most models do not perform well in terms of encoding length in the latent space as recorded by the discrepancy along the $T(\tilde X_*|X_*)$ dimension, with the exception of high R models that seem to encode it to some extent. Models with low R (and collapsed posteriors) model length unconditionally as you might expect from large transformer based language models. 
The {\bf PTB latent topics} paint a rather different picture. The tendency for high R and the conditional statistic $T(\tilde X_*|X_*)$ is flipped: higher R only harms along this axis and correspondingly along the other statistical axes. 

For both MNIST and PTB, \infovae with high enough $\lambda_{\text{MMD}}$ weight can diminish divergence from the prior as measured by the latent analysis model. This does not, however, translate in consistent effects in the data space.

\section{Conclusion}\label{sec:discussion}

We have demonstrated the use of Bayesian data analysis techniques to criticise VAE in terms of the two probability distributions they prescribe. By employing LSMs that can be purpose chosen by a practitioner, we probe the VAEs for distributional  compliance relative to a relevant control group. The lppd statistic is low-dimensional but reflects the LSM's richness, besides that it naturally accommodates a marginal and a conditional view, both crucial for analysing VAEs. We show that this methodology offers possibilities for model selection qualitatively beyond intrinsic evaluation metrics and at finer granularity than commonly used statistics can offer. Unlike NHST procedures, the components of our methodology are themselves amenable to model checking techniques, which help us build trust in the analysis.
As our methodology interacts with the LVM only as a sampler, it can be extended to other LVMs such as GANs \citep{goodfellow2014generative} and diffusion models \citep{sohl2015deep,kingma2021variational}.

\clearpage  %

\begin{acknowledgements} %
    This project has received funding from the European Union's Horizon 2020 research and innovation programme under grant agreement No 825299  (GoURMET).
    
\end{acknowledgements}

\bibliography{bib}

\appendix

\providecommand{\upGamma}{\Gamma}
\providecommand{\uppi}{\pi}

\clearpage %
\onecolumn
\section{Acknowledgements to open-source software projects}

We would like to give credit to some essential software projects we used extensively for our work. We would like to at least name Pyro \citep{bingham2019pyro}, NumPyro \citep{phan2019numpyro}, Numpy \citep{harris2020numpy}, PyTorch \citep{pytorch2019} and Scikit Learn \citep{scikit} as essential tool boxes for the experiments we have conducted and the analyses we have performed.

\section{Experimental set-up}\label{app:hyperparameters}
\subsection{Objectives}

An overview of the objectives used in the experiments as described in Section \ref{sec:experiments} is given in Table \ref{tab:objectives}. The hyperparameter settings of the separate runs are summarised in Table \ref{tab:objectives-hp}. The total number of experiments ran on the Penn Treebank dataset is 23 and on binarised MNIST 66.

\begin{table*}[!htb]
    \centering
    \scriptsize
    \begin{tabular}{ccc}
        \toprule
        Objective (hyperparameters) & Equation \\
        \midrule
        $\beta$-VAE \cite{higgins2016beta} ($\beta$) & $\max_{\theta, \phi} \mathcal{L}_{\beta\text{-VAE}}(\theta, \phi) = \mathbb{E}_{q_{Z|X=x}}[\log p_{X|Z=z}(\phi)] - \beta \KL(q_{Z|X=x}(\theta)||p_{Z}(\phi))$ \\
        \addlinespace[0.5em]
        Info-VAE \cite{zhao2017infovae} ($\lambda_\text{rate}$, $\lambda_\text{MMD}$) & $\max_{\theta, \phi} \mathcal{L}_{\text{Info-VAE}}(\theta, \phi) = \mathbb{E}_{q_{Z|X=x}}[\log p_{X|Z=z}(\phi)] - \lambda_\text{rate} \KL(q_{Z|X=x}(\theta)||p_{Z}(\phi)) + \lambda_\text{MMD} \text{MMD}$\\
        \addlinespace[0.5em]
        Free-bits-VAE \cite{kingma2016improved}  ($\lambda_\text{FB}$) & $\max_{\theta, \phi} \mathcal{L}_{\text{FB-VAE}}(\theta, \phi) = \mathbb{E}_{q_{Z|X=x}}[\log p_{X|Z=z}(\phi)] - \max( \KL(q_{Z|X=x}(\theta)||p_{Z}(\phi)), \lambda_{\text{FB}})$ \\
        \bottomrule
    \end{tabular}
    \caption{Overview of the objectives with their hyperparameters used for the experiments.}
    \label{tab:objectives}
\end{table*}

\begin{table*}[!htb]
    \centering
    \begin{tabular}{c||c|c}
        \toprule
         & PTB & MNIST \\
         \midrule
        $\beta \in$ & $\{0.0, 0.25, 0.5, 0.75, 1.0, 2.0\}$ & $\{0.0, 0.25, 0.5, 0.75, 1.0, 1.5, 2.0, 5.0, 10.0\}$ \\
        \addlinespace[0.5em]
        $(\lambda_{\text{rate}}, \lambda_{\text{MMD}}) \in$ & $\{1, 10, 100\} \times \{0.1, 0.5, 1.0, 2.0\} $ & $\{1, 10, 100\} \times \{0.1, 0.5, 1.0, 2.0, 5.0, 10.0\}$ \\
        \addlinespace[0.5em]
        $\lambda_{\text{FB}} \in$ & $\{4, 8, 16, 32, 64\}$ & $\{4, 8, 16, 24, 32, 40\}$ \\
        \bottomrule
    \end{tabular}
    \caption{The hyperparameters for the objectives outlined in Section \ref{sec:experiments}}.
    \label{tab:objectives-hp}
\end{table*}

\section{Architectures}\label{app:architectures}

\subsection{Binarised MNIST}
\subsubsection{Gated CNN Encoder}

We use the the gated convolutional encoder from \citet{van2018sylvester} with two additional linear layers to map to the location and scale parameters of the approximate posterior. The gating mechanism can be expressed as follows, where $\ast$ denotes convolution and $\odot$ denotes element-wise multiplication:

\begin{equation*}
    \mathbf{y}_{\text{GatedConv2D}} = (\mathbf{V} \ast \mathbf{x} + \mathbf{b}) \odot \sigma(\mathbf{W} \ast \mathbf{x} + \mathbf{c})
\end{equation*}

The encoder consists of the following GatedConv2d layers, with the parameters between parentheses denoting number of input channels, number of output channels, kernel size, stride and padding respectively:

\begin{itemize}
    \item GatedConv2d(1,  32,  5, 1, 2)
    \item GatedConv2d(32, 32,  5, 2, 2))
    \item GatedConv2d(32, 64,  5, 1, 2)
    \item GatedConv2d(64, 64,  5, 2, 2)
    \item GatedConv2d(64, 64,  5, 1, 2)
    \item GatedConv2d(64, 256, 7, 1, 0)
\end{itemize}

\subsubsection{Gated CNN.T decoder}

Similarly, for the simple decoder architecture we follow \citet{van2018sylvester} by using GatedConvTranspose2d units as the main building block to map the sampled latent representation $\mathbf{z}$ to the parameters of Bernoulli distributions to model the binary pixel values. The full architecture can be summarised as follows, where the parameters of the GatedConvTranspose2d, in order, denote the number of input channels, the number of output channels, kernel size, stride, padding and (optionally) the output padding:

\begin{itemize}
    \item GatedConvTranspose2d(10, 64, 7, 1, 0)
    \item GatedConvTranspose2d(64, 64, 5, 1, 2)
    \item GatedConvTranspose2d(64, 32, 5, 2, 2, 1)
    \item GatedConvTranspose2d(32, 32, 5, 1, 2)
    \item GatedConvTranspose2d(32, 32, 5, 2, 2, 1)
    \item GatedConvTranspose2d(32, 32, 5, 1, 2)
    \item GatedConvTranspose2d(32, 1, 1, 1, 0)
\end{itemize}

\subsubsection{PixelCNN++ decoder}

We follow \cite{alemi2018fixing} in slightly modifying the work of \cite{salimans2017pixelcnn++} to function as a VAE decoder architecture. The latent $\mathbf{z}$ is added to the decoder via a conditioning mechanism in all the GatedResNet blocks. This mechanism projects the latent to the spatial dimensions of the feature maps that are the output of this block ($\mathbf{x_1}$ and $\mathbf{x_2}$) and and adds the projections to all channels identically before the gating mechanism. The operation can be described as:

\begin{equation*}
    \mathbf{y_{\text{GatedResNet}}} = (\mathbf{x_1} + \mathbf{V}^T \mathbf{z}) \odot \sigma(\mathbf{x_2} + \mathbf{W}^T \mathbf{z})
\end{equation*}

We use three down-sampling blocks ($28 \rightarrow 14 \rightarrow 7$) and three up-sampling blocks ($7 \rightarrow 14 \rightarrow 28$) with skip connections between blocks of equal spatial dimensionality. Each block consists of 2 GatedResNet units and the number of filters is set to 64. We adapt the output layer to output the parameters of the Bernoulli distribution over the spatial dimensions of the image (28 x 28).

\subsection{Penn Treebank}
\subsubsection{Distil RoBERTa Encoder} 

For the encoder we use a transformer architecture, specifically a RoBERTa Encoder architecture \cite{liu2019roberta}\footnote{We use the implementation of Huggingface described here: \url{https://huggingface.co/docs/transformers/model_doc/roberta}} and initialise with weights that are obtained by means of knowledge distillation \cite{Sanh2019DistilBERTAD}.\footnote{Specifically we use the weights from the \texttt{distil-roberta} checkpoint: \url{https://huggingface.co/distilroberta-base}} We add a pooling layer that maps the output of the encoder to the parameters of the approximate posterior distribution.

\subsubsection{Adapted Distil Roberta Decoder} 
For the decoder we use the same basis as for the encoder. We adapt its architecture following the work of \cite{li2020optimus} to incorporate the latent representations via two mechanisms: the attention mechanism and the embedding mechanism. For the former, the latent representation is mapped to the dimensionality of the hidden layers for every layer in the model and added via the attention mechanism in the form of key and value vectors. For the latter mechanism, the latent representation is simply projected to the dimensionality of the hidden layers and is summed with the initial hidden states of the model, right after embedding the tokens. This operation is the same for all positions in the sequence. At the output, a language model head is added to map the output of the RoBERTa block to the parameters of a Categorical distribution that models the token distributions per sequence position. Auto-regressive masking is used to prevent the model to have access to information beyond the current token position.

\section{Latent structure models and lppd statistics}\label{app:latent-structure-models}
\subsection{Graphical models}

Table \ref{tab:BDA_diagrams} shows the graphical models of the latent structure models used for our analysis.

\begin{table*}[!htb]
    \centering
    \small
    \begin{tabular}{ ccccc } 
    \toprule
    MNIST & \multicolumn{2}{c}{PTB} & Latent  \\
    
    \cmidrule(lr){1-1}
    \cmidrule(lr){2-3}
    \cmidrule(lr){4-4} \\
    
    \begin{tikzpicture}[x=1.3cm, y=0.8cm]
      
        \node[obs]                   (x_n)    {$x_n$} ; %
        \node[latent, above=of x_n, yshift=+0.5cm]  (p_kd)   {$p_{k d}$} ; %
        \node[latent, left=of x_n]   (c_n)    {$c_n$} ;
        
        \node[const, above=of p_kd, xshift=-0.5cm]  (beta_p) {$\alpha_p$}; 
        \node[const, above=of p_kd, xshift=+0.5cm]  (alpha_p) {$\beta_p$};

        {\tikzset{plate caption/.append style={below=0.15cm of #1.south east}}
        \plate[minimum width=1.45cm, minimum height=1.55cm, yshift=+0.1cm, xshift=-0.05cm]{component_plate}{(p_kd)}{\footnotesize $K$};}
        
        {\tikzset{plate caption/.append style={below=0.65cm of #1.south east, xshift=+0.05cm}}
        \plate[minimum width=1.85cm, xshift=-0.1cm, minimum height=4.25cm, yshift=+0.1cm]{pixel_plate}{(p_kd)(x_n)}{\footnotesize $D$};}
        
        {\tikzset{plate caption/.append style={below=0.1cm of #1.south east, xshift=+0.1cm}}
        \plate[minimum height=1.6cm, yshift=+0.15cm]{data_plate}{(c_n)(x_n)}
        {\footnotesize $N$};}

        \edge{p_kd}{x_n}
        \edge{alpha_p, beta_p}{p_kd}
        \edge{c_n}{x_n}
      
    \end{tikzpicture}

    & 
    \begin{tikzpicture}[x=1.2cm, y=0.8cm]
      \node[obs]                   (x_n)            {$x_n$};
      \node[latent, above=of x_n]  (c_n)            {$c_n$};
      \node[latent, above=of c_n]  (theta)          {$\vec{\theta}$};
      \node[const, above=of theta] (alpha_theta)    {$\alpha_\theta$};
    
      \node[latent, left=of c_n, xshift=0.35cm] (rate_k)  {$r_k$}; %
      \node[const, above=of rate_k, xshift=-0.5cm]  (a_rate) {$a_\phi$}; 
      \node[const, above=of rate_k, xshift=+0.5cm]  (b_rate) {$b_\phi$}; %
    
      \plate[minimum height=2.8cm, minimum width=1.35cm]{plate_obs} {(c_n) (x_n) } {$N$};
      \plate[minimum height=1.55cm, yshift=+0.1cm, minimum width=1.35cm]{plate_rate} {(rate_k)} {$K$}
      
      \edge {c_n,rate_k} {x_n} ; %
      \edge {theta} {c_n}
      \edge {alpha_theta} {theta}
      \edge {a_rate}{rate_k}
      \edge {b_rate}{rate_k}
    \end{tikzpicture} 
    
    & 
    \begin{tikzpicture}[x=1.3cm, y=0.8cm]
        
        \node[const] (beta) {$\beta_\phi$};
        \node[latent, above=of beta, yshift=+0.25cm] (phi_k) {$\vec{\phi}_k$};
        \node[obs, above=of phi_k, yshift=0.8cm] (x_mn) {$x_{mn}$};
        \node[latent, above=of x_mn] (c_mn) {$c_{mn}$};
        \node[latent, above=of c_mn] (theta_m) {$\vec{\theta}_m$};
        \node[const, above=of theta_m] (alpha) {$\alpha_\theta$};

        {\tikzset{plate caption/.append style={below=0.85cm of #1.south east}}
        \plate[minimum width=2.1cm, minimum height=5.4cm, yshift=+0.1cm, xshift=0.0cm]{document_plate}{(theta_m)(c_mn)(x_mn)}{$M$}}
        
        {\tikzset{plate caption/.append style={below=0.3cm of #1.south east}}
        \plate[minimum width=1.6cm, xshift=+0.0cm, yshift=0.1cm, minimum height=3.4cm]{token_plate}{(c_mn)(x_mn)}{$N$}}
        
        {\tikzset{plate caption/.append style={below=0.4cm of #1.south east}}
        \plate[minimum width=2.1cm, minimum height=1.6cm, yshift=-0.05cm]{topic_plate}{(phi_k)}{$K$}}
        
        \edge{alpha}{theta_m}
        \edge{theta_m}{c_mn}
        \edge{c_mn}{x_mn}
        \edge{phi_k}{x_mn}
        \edge{beta}{phi_k}
  
    \end{tikzpicture}
    
    &
    \begin{tikzpicture}[x=1.3cm, y=0.8cm]

        \node[latent] (d) {$\vec{d}$};
        
        \node[latent, above=of d, yshift=+0.25cm] (lambda_k) {$\vec{\lambda}_k$} ;
        
        \node[latent, below=of lambda_k, yshift=-0.25cm, xshift=+1.0cm] (mu) {$\vec{\mu}$} ;
        \node[latent, below=of lambda_k, yshift=-0.25cm, xshift=-1.0cm] (f) {$\vec{f}$} ;
        
        \node[obs, above=of lambda_k, yshift=0.8cm] (z_mn) {$\vec{z}_{mn}$} ;
        \node[latent, above=of z_mn] (c_mn) {$c_{mn}$} ;
        \node[latent, above=of c_mn] (theta_m)  {$\vec{\theta}_m$} ;
        \node[const, above=of theta_m] (alpha)    {$\alpha_\theta$} ;
        
        \node[const, below= of mu] (mu_prior) {$\{ \alpha_\mu \}$};
        \node[const, below= of f] (f_prior) {$\{ \alpha_f \}$};
        \node[const, below= of d] (d_prior) {$\{ \alpha_d \}$};

        {\tikzset{plate caption/.append style={below = 0.85cm of #1.south east}}
        \plate[minimum width=2.1cm, minimum height=5.4cm, yshift=+0.1cm, xshift=0.0cm]{document_plate}{(theta_m)(c_mn)(z_mn)}{$M$}}
        
        {\tikzset{plate caption/.append style={below = 0.45cm of #1.south east}}
        \plate[minimum width=1.6cm, xshift=+0.0cm, yshift=0.1cm, minimum height=3.4cm]{token_plate}{(c_mn)(z_mn)}{$N$}}
        
        {\tikzset{plate caption/.append style={above right = -0.15cm and 0.15cm of #1.north east}}
        \plate[minimum width=2.0cm, minimum height=1.5cm, yshift=-0.25cm, xshift=-0.15cm]{topic_plate}{(lambda_k)}{$K$}}
        
        \edge{alpha}{theta_m}
        \edge{theta_m}{c_mn}
        \edge{c_mn}{z_mn}
        \edge{lambda_k}{z_mn}
        \edge{mu}{lambda_k}
        \edge{d}{lambda_k}
        \edge{f}{lambda_k}
        
        \edge{mu_prior}{mu}
        \edge{d_prior}{d}
        \edge{f_prior}{f}
  
    \end{tikzpicture}
    \\
    \\
    Digit identity & Sequence length & Topic structure & Prior structure \\
    \bottomrule
    \end{tabular}
    \caption{This table shows the latent structure models described in Section \ref{subsec:bda_models} used to demonstrate the proposed evaluation methodology. The captions denote the latent structure that is captured by each individual model and is represented graphically by the latent variable $c$.}
    \label{tab:BDA_diagrams}
 \end{table*}

\subsection{Model checks}

In the following sections we will provide material to assess goodness of fit of the latent structure models used in our analysis.

\subsubsection{MNIST digit identity}

\begin{figure}[h!]
    \centering
    \includegraphics[width=0.5\textwidth]{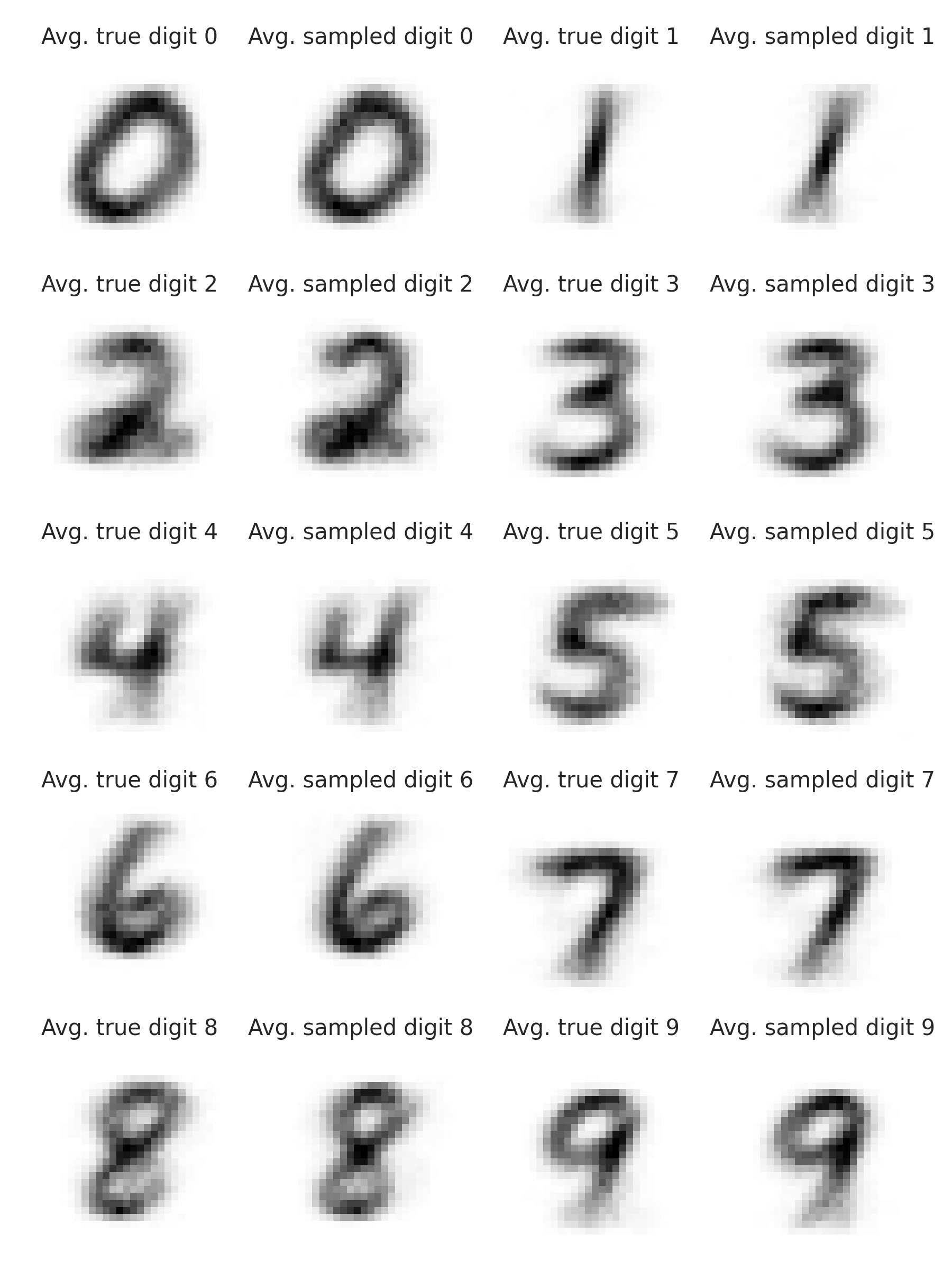}
    \caption{Average sampled digits from the held-out dataset are plotted next (left) to average sampled digits from the posterior predictive of the MNIST digit identity latent structure model (right).}
    \label{fig:bda_check_mnist_sample_plot}
\end{figure}

\begin{figure*}[h!]
    \centering
    \includegraphics[width=\textwidth]{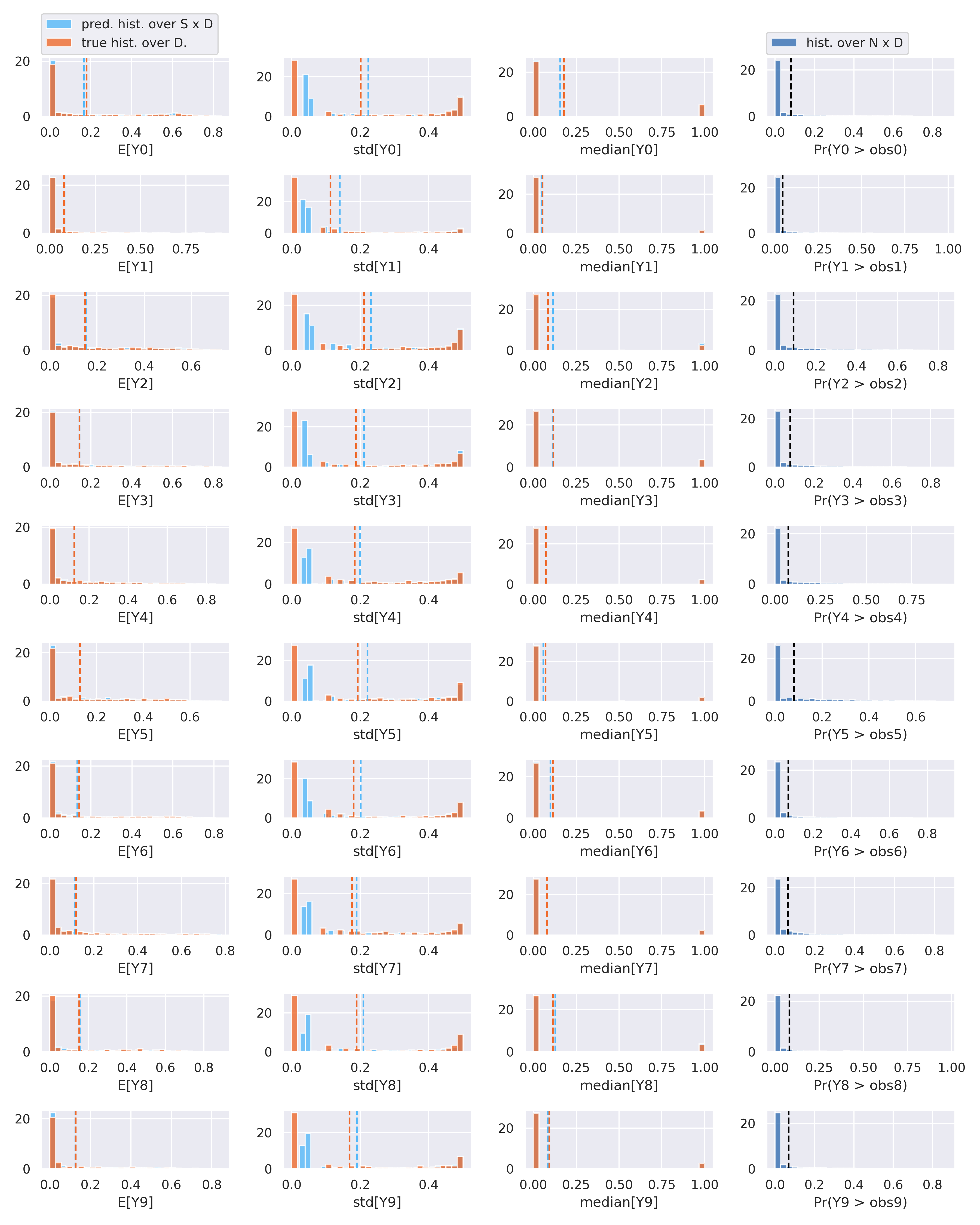}
    \caption{Posterior predictive checks for the MNIST digit identity latent structure model fit to the data, row-wise organised per digit group. From left to right it shows histograms over the mean, the standard deviation, the median and the probability that a sampled pixel value from the posterior predictive exceeds that of a sampled pixel from held out observations.}
    \label{fig:bda_check_mnist_posterior_predictive_checks}
\end{figure*}

In Figure \ref{fig:bda_check_mnist_sample_plot} we show average digits sampled from the held out dataset next to average digits sampled from the posterior predictive of the MNIST digit identity model as presented in section \ref{sec:experiments} to assess its fit. In Figure \ref{fig:bda_check_mnist_posterior_predictive_checks} numerical posterior predictive checks are shown.

\subsubsection{PTB sequence length}

\begin{figure*}[!htb]
    \centering
    \includegraphics[width=0.7\textwidth]{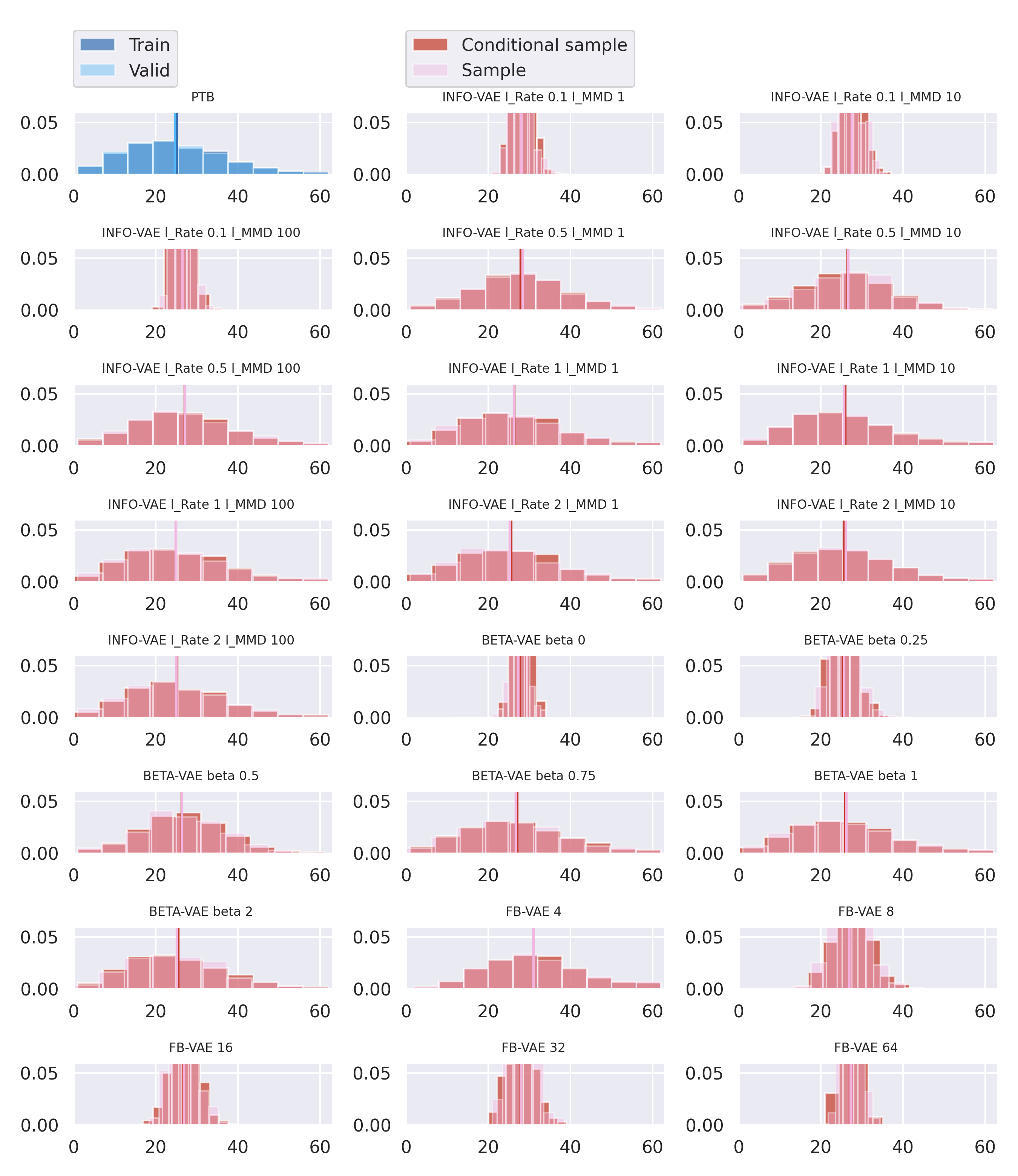}
    \caption{Sequence length histograms of model samples (pink and red hues) and data samples (blue hues).}
    \label{fig:ptb_data_model_length_distributions}
\end{figure*}

\begin{figure*}[!htb]
    \centering
    \includegraphics[width=\textwidth]{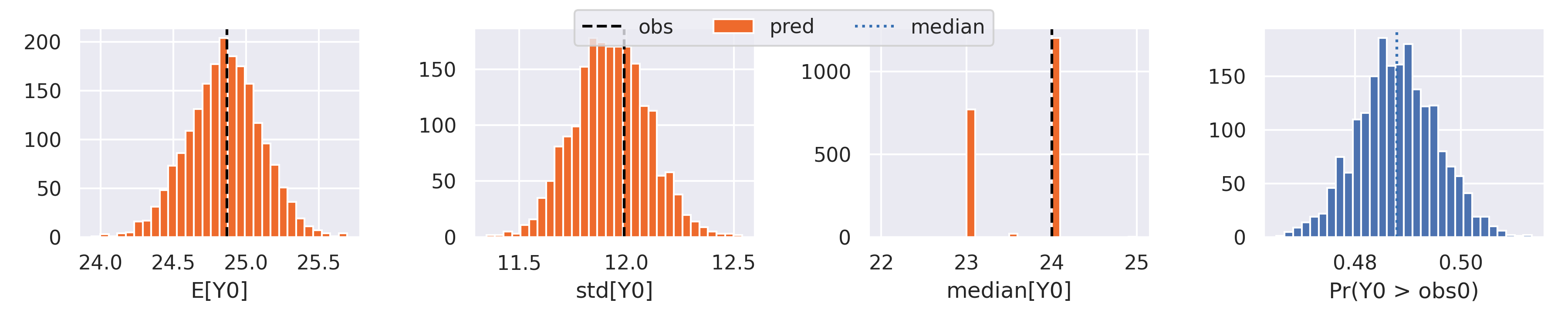}
    \caption{Posterior predictive checks for the sequence length latent structure model. From left to right it shows histograms over the mean, the standard deviation, the median and the probability that a sampled length value from the posterior predictive exceeds that of a sampled length from held out observations.}
    \label{fig:bda_check_ptb_seq_len_posterior_predictive_checks_1}
\end{figure*}

\begin{figure*}[!htb]
    \centering
    \includegraphics[width=0.6\textwidth]{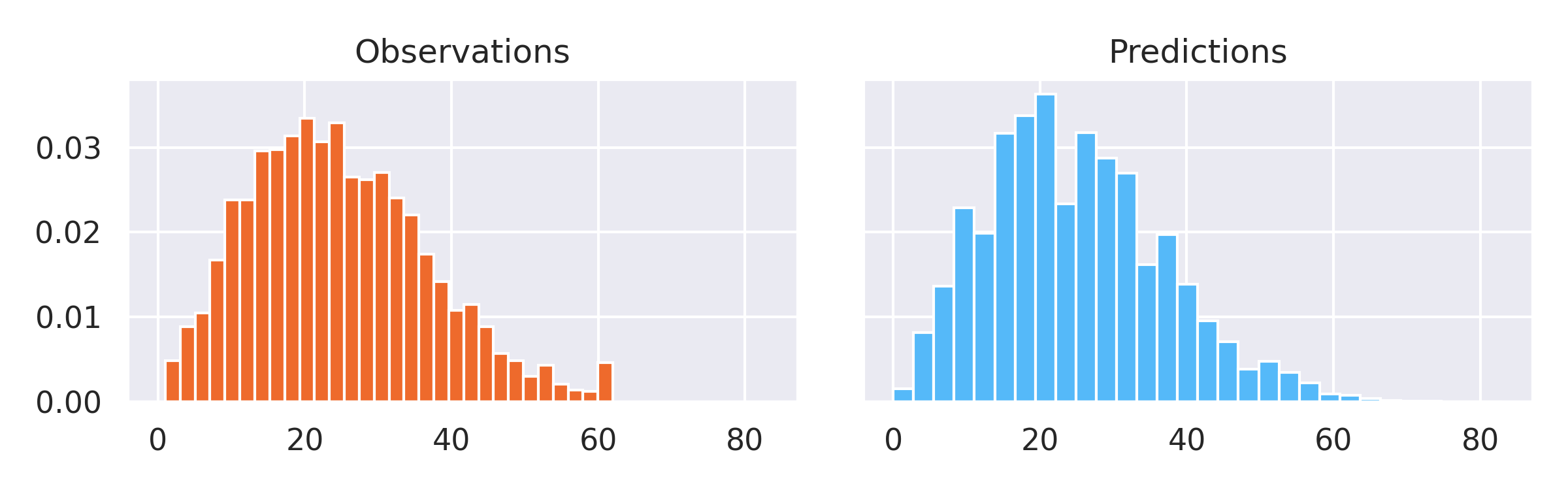}
    \caption{A histogram of sampled lengths from the held out Penn Treebank dataset (left) next to sampled lengths from the posterior predictive of the sequence length latent structure model (right).}
    \label{fig:bda_check_ptb_seq_len_posterior_predictive_checks_2}
\end{figure*}

In Figure \ref{fig:ptb_data_model_length_distributions} histograms of the sampled lengths of the models trained on the Penn Treebank dataset are shown together with the true lengths. This is the raw data used for the sequence length latent structure model. Posterior predictive checks for this latent structure model fit on the held out data are shown in Figure \ref{fig:bda_check_ptb_seq_len_posterior_predictive_checks_1} and Figure  \ref{fig:bda_check_ptb_seq_len_posterior_predictive_checks_2}.

\subsubsection{PTB topics}

\begin{figure*}[!htb]
    \centering
    \includegraphics[width=0.75\textwidth]{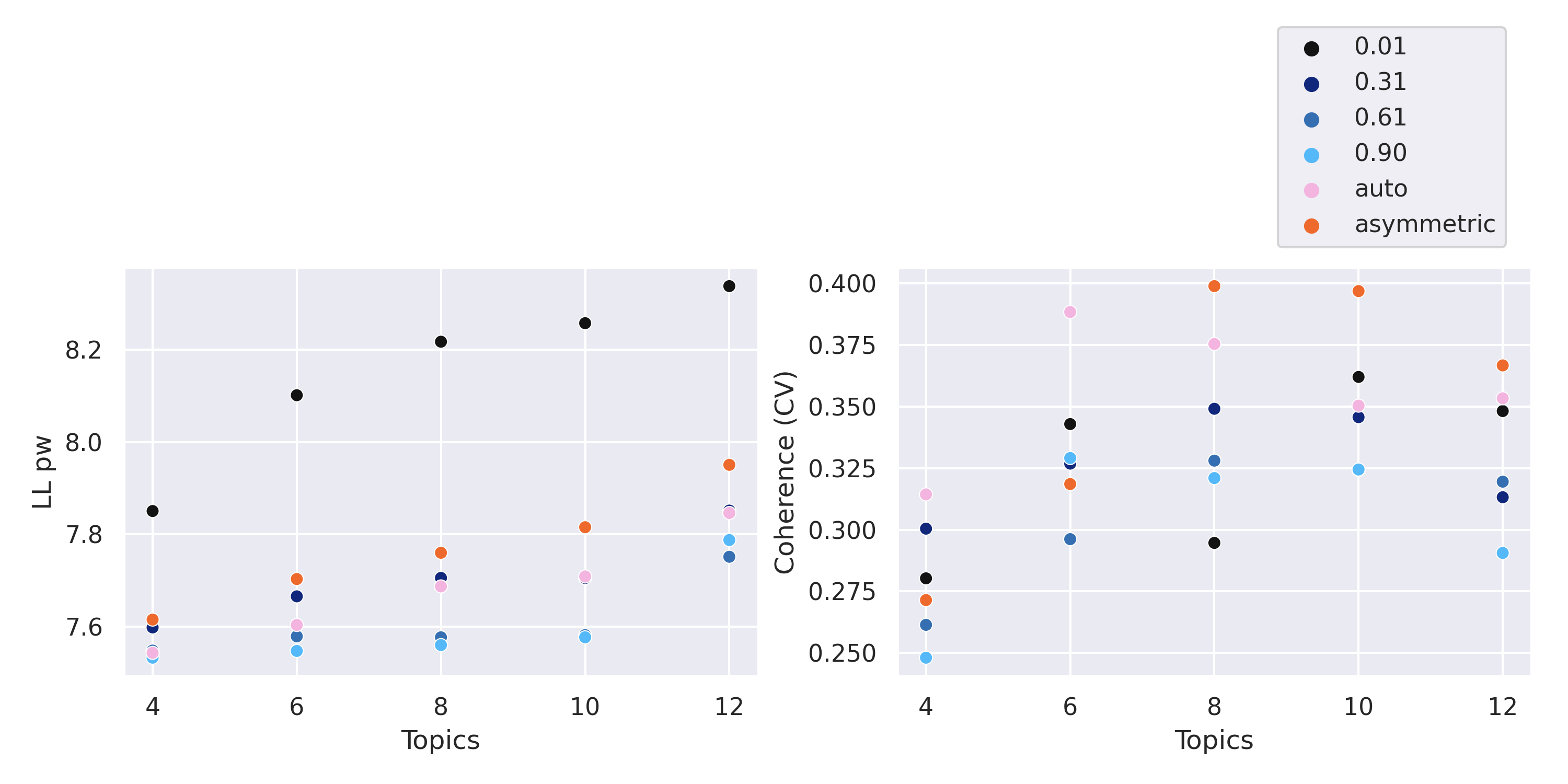}
    \caption{The per-token log likelihood bound and a sliding window topic coherence for different number of topics and different values for \texttt{alpha}.}
    \label{fig:bda_checks_lda_hp_tuning}
\end{figure*}

\begin{figure*}[!htb]
    \centering
    \includegraphics[width=0.8\textwidth]{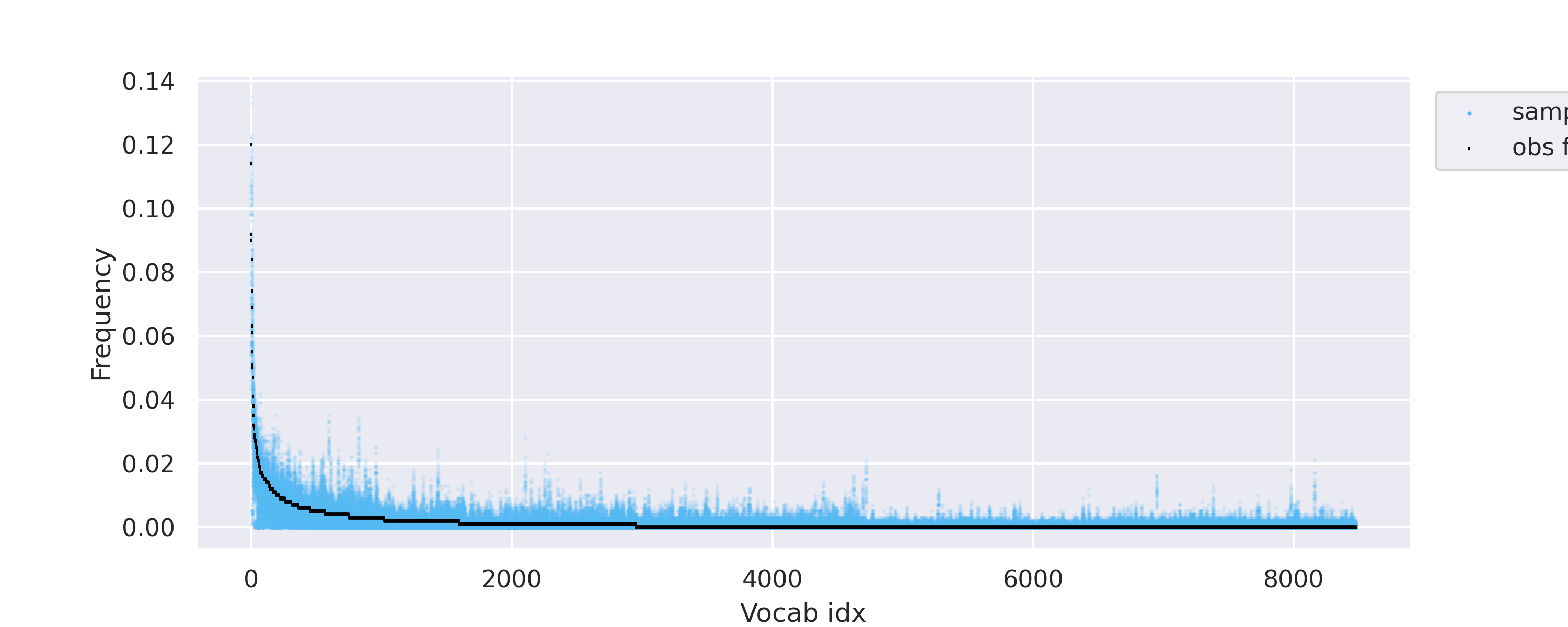}
    \caption{True sorted token frequencies of the Penn Treebank data set plotted against re-sampled sorted token frequencies sampled from the fitted LDA model.}
    \label{fig:bda_checks_lda_word_freqs}
\end{figure*}

\begin{table*}[!htb]
    \centering
    \scriptsize
    \begin{tabular}{rllllllllll}
\toprule
 Topic &  Token 0 & Token 1 &     Token 2 &    Token 3 &    Token 4 &    Token 5 &   Token 6 &   Token 7 &         Token 8 &    Token 9 \\
\midrule
     0 &   market &   stock &       price &    trading &   investor &        day &    trader &    future &           index &        buy \\
     1 &       mr &     say &         big &        dow &      jones &  dow\_jones &       one &     point &           world &      going \\
     2 &     test &      mr &      cancer &      house &  treatment &   director &     state &   federal &         whether &     senate \\
     3 &  million &   share &        year &    billion &       sale &    earlier &   quarter &   company &            rose &      month \\
     4 &  company &     new &  laboratory &  operation &    billion &       corp &     calif &      loan &             los &    deficit \\
     5 &     bond &    rate &        year &     dollar &     market &      franc &      bank &  interest &         billion &      yield \\
     6 &  company &     inc &       third &       corp &  president &  executive &     chief &        co &   third\_quarter &     unilab \\
     7 &      new &    york &    new\_york &      stock &   exchange &      month &    closed &   trading &  stock\_exchange &  yesterday \\
     8 &     next &    data &     control &     export &      would &    company &  spending &      year &        computer &     friday \\
     9 &       mr &     say &        year &      could &      buyer &    central &      drug &       old &            time &      would \\
\bottomrule
\end{tabular}
    \caption{The top 10 tokens per topic as identified by the LDA topic model.}
    \label{tab:lda_topic_words}
\end{table*}

To choose the hyperparameters of the LDA topic model that is fit to the Penn Treebank data, we perform a hyperparameter experiment. To this end, we compute the per-token log likelihood bound and a sliding window topic coherence score (Gensim's \texttt{c\_v}), using the held-out Penn Treebank data. Based on these results (visualised in Figure \ref{fig:bda_checks_lda_hp_tuning}) we set \texttt{alpha} to 0.01 and set the number of topics to 10. Figures \ref{fig:bda_checks_lda_word_freqs} shows an additional check to assess goodness of fit for the LDA model given these hyperparameters. %

\subsubsection{Latent space}\label{app:latent-space}

\begin{figure*}[!htb]
    \centering
    \includegraphics[width=0.8\textwidth]{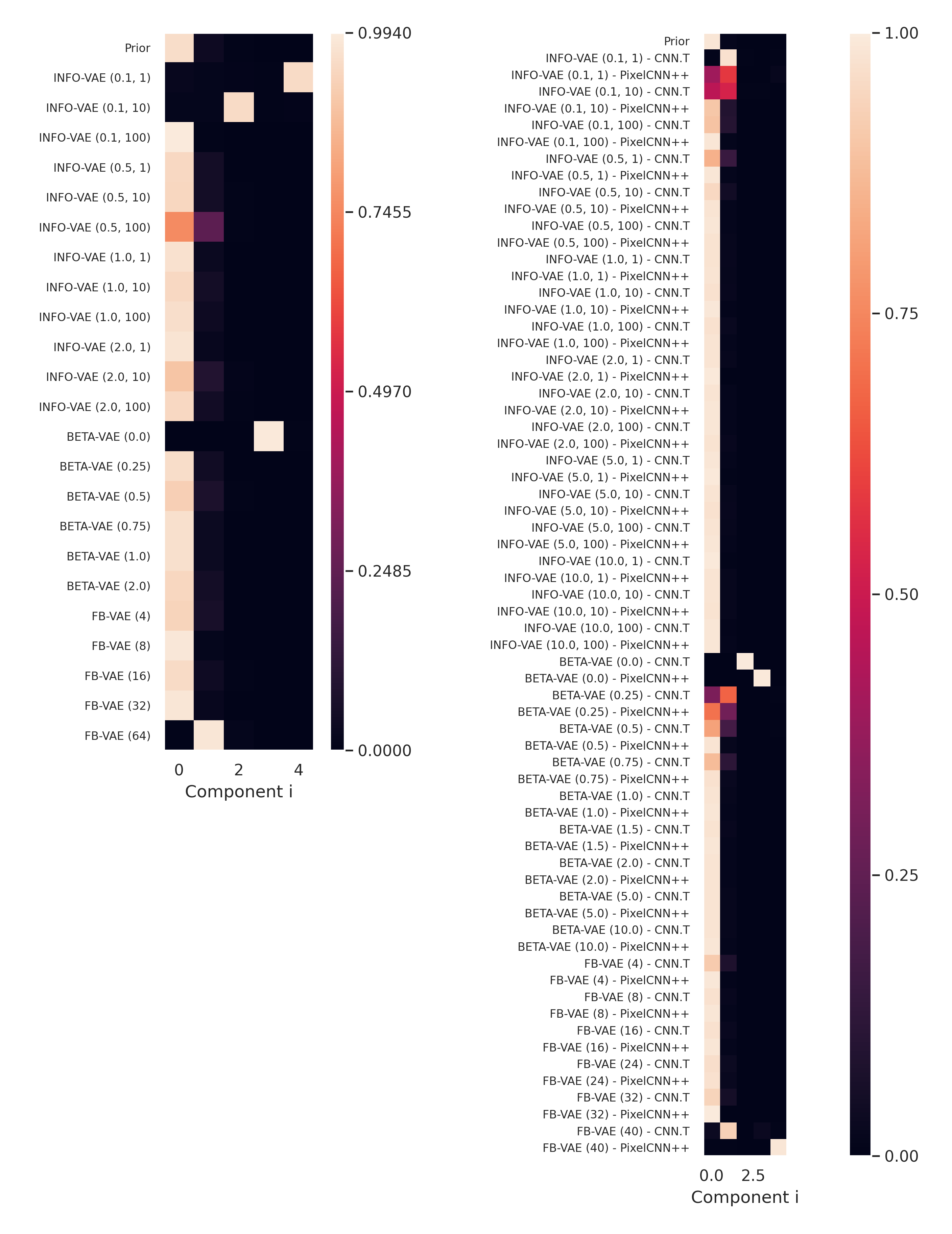}
    \caption{Sampled components from the posterior of the latent analysis models for all experiment groups and the control (data group). The left column shows the PTB experiments, right the MNIST experiments.}
    \label{fig:latent-component-plots}
\end{figure*}

In Figure \ref{fig:latent-component-plots} we plot the average posterior component sample for the latent analysis models.

\subsection{Full experimental results lppd statistics}

\begin{figure*}[!htb]
    \centering
    \includegraphics[width=\textwidth]{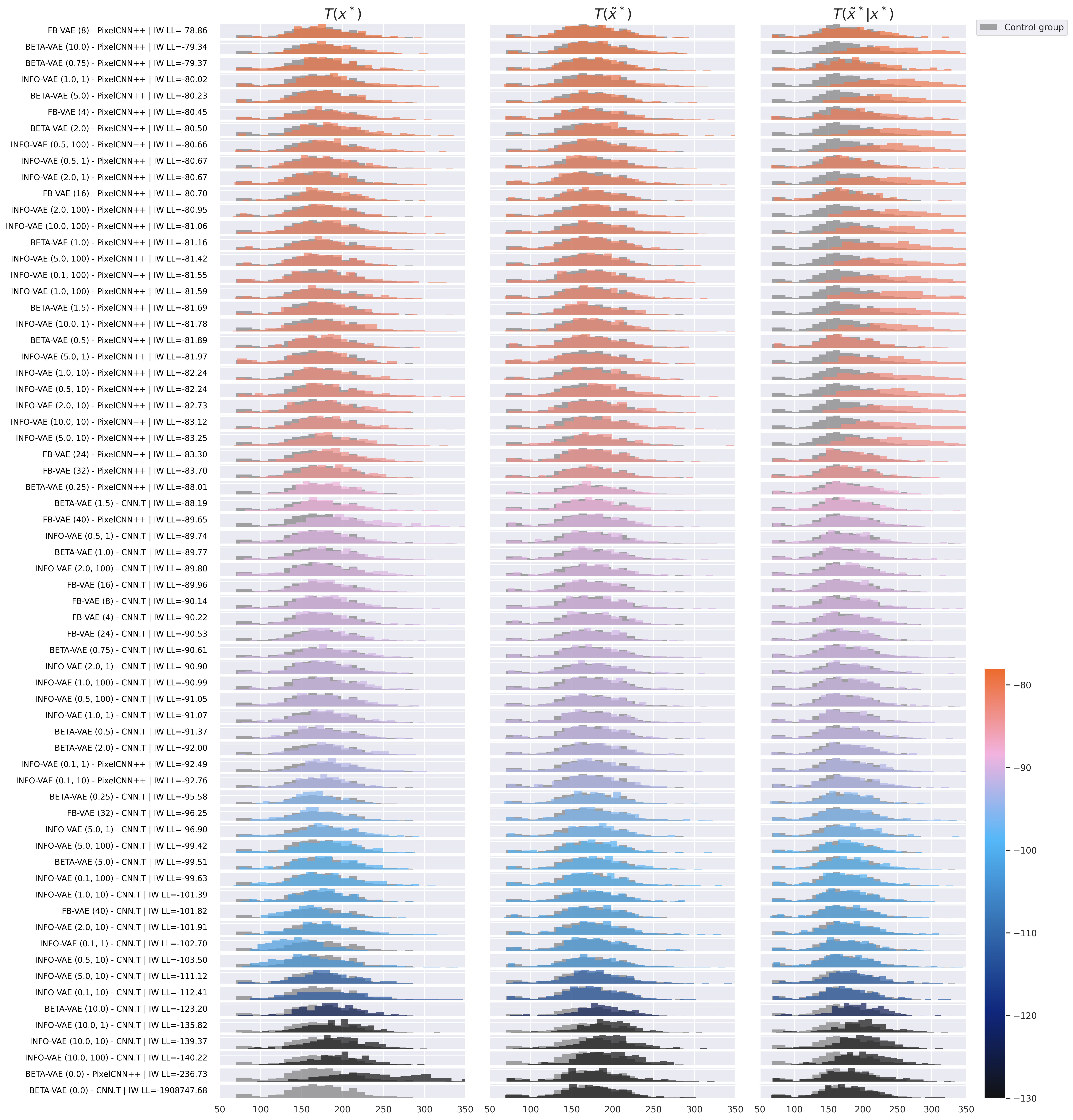}
    \caption{The three statistics $T(x^*)$, $T(\tilde x^*)$ and $T(\tilde x^*|x^*)$ based on the log posterior predictive density of the MNIST digit identity latent structure model of the experiments plotted against the respective control groups. The rows are ordered and coloured by an importance weighted estimate of the log likelihood (IW LL) of the VAEs on held-out data.}
    \label{fig:mnist_surprisal_dist}
\end{figure*}

\begin{figure*}[!htb]
    \centering
    \includegraphics[width=\textwidth]{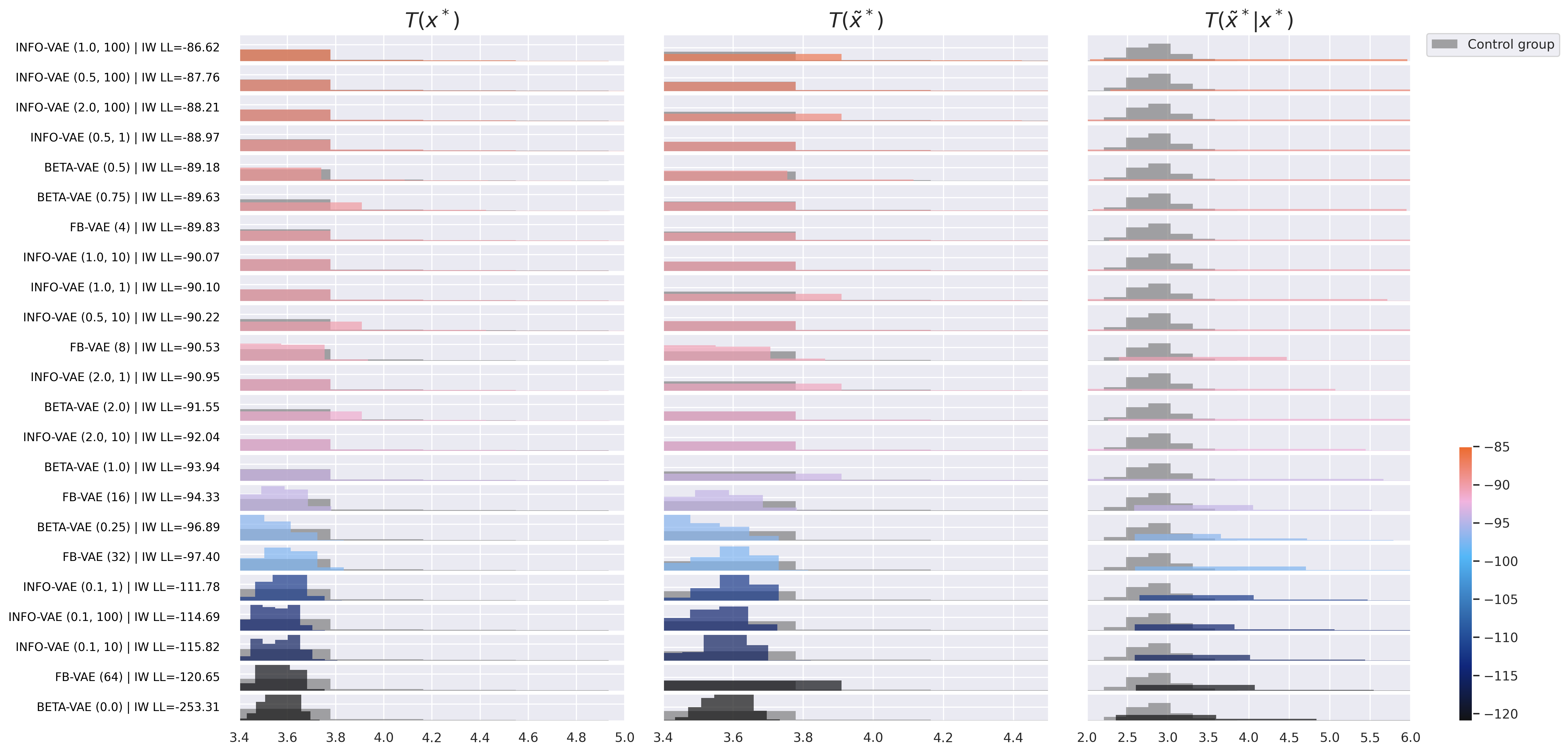}
    \caption{The three statistics $T(x^*)$, $T(\tilde x^*)$ and $T(\tilde x^*|x^*)$ based on the log posterior predictive density of the PTB sequence length latent structure model of the experiments plotted against the respective control groups. The rows are ordered and coloured by an importance weighted estimate of the log likelihood (IW LL) of the VAEs on held-out data.}
    \label{fig:ptb_seq_len_surprisal_dist}
\end{figure*}

\begin{figure*}[!htb]
    \centering
    \includegraphics[width=\textwidth]{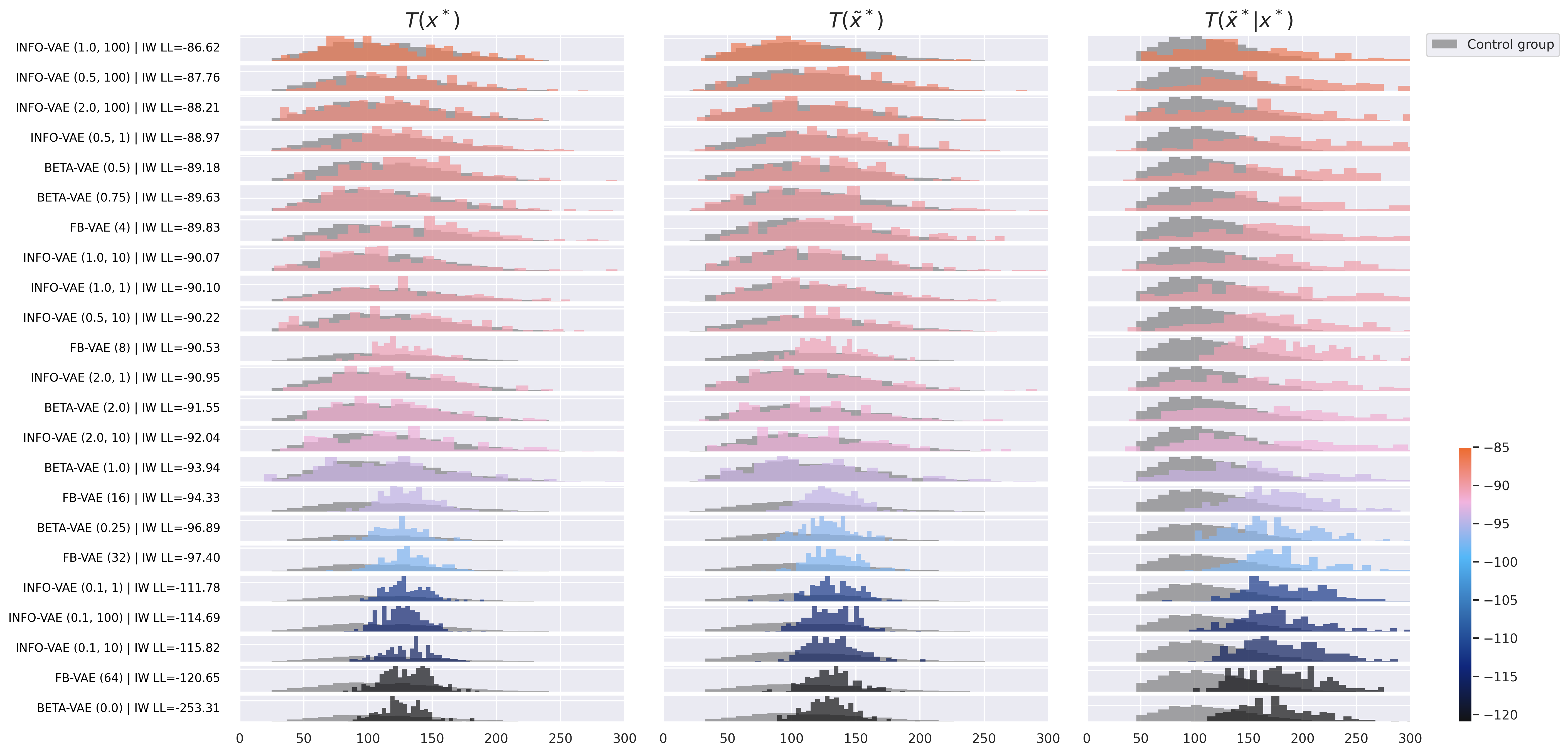}
    \caption{The three statistics $T(x^*)$, $T(\tilde x^*)$ and $T(\tilde x^*|x^*)$ based on the log posterior predictive density of the PTB topic latent structure model of the experiments plotted against the respective control groups. The rows are ordered and coloured by an importance weighted estimate of the log likelihood (IW LL) of the VAEs on held-out data.}
    \label{fig:ptb_lda_topics_surprisal_dist}
\end{figure*}

Full experimental results for the three type of statistics measured under the three latent structure models can be found in Figures \ref{fig:mnist_surprisal_dist}, \ref{fig:ptb_seq_len_surprisal_dist} and \ref{fig:ptb_lda_topics_surprisal_dist}.

\section{Intrinsic evaluation results}\label{app:intrinsic-evaluation}

The full results as visualised in Figure \ref{fig:intrinsic-evaluation-plot} are listed in Tables \ref{tab:bmnist_full_results} and \ref{tab:ptb_full_results} for the binarised MNIST and PTB experiments respectively.

\begin{table*}[!htb]
    \centering
    \scriptsize
    \begin{tabular}{llllll|rrrrr}
    \toprule
    Decoder & Objective & $\lambda_{\text{FB}}$ &  $\beta$ & $\lambda_{\text{MMD}}$ & $\lambda_{\text{rate}}$ &    IW LL &     ELBO &    Rate &  Distortion &        MMD \\
\midrule
      CNN.T &  BETA-VAE &         - &  0.00 &     - &      - & -1908747.684 & -1910785.802 & 1910706.699 &      79.108 &   0.265494 \\
 PixelCNN++ &  BETA-VAE &         - &  0.00 &     - &      - &     -236.732 &     -235.079 &     174.957 &      60.122 &   0.530988 \\
      CNN.T &  INFO-VAE &         - &      - &   100 & 10.0 &     -140.217 &     -157.334 &       3.218 &     154.116 &   0.000446 \\
      CNN.T &  INFO-VAE &         - &      - &    10 & 10.0 &     -139.367 &     -157.106 &       3.275 &     153.831 &   0.000721 \\
      CNN.T &  INFO-VAE &         - &      - &     1 & 10.0 &     -135.821 &     -155.290 &       3.412 &     151.878 &   0.000594 \\
      CNN.T &  BETA-VAE &         - & 10.00 &     - &      - &     -123.197 &     -142.718 &       4.104 &     138.614 &   0.000462 \\
      CNN.T &  INFO-VAE &         - &      - &    10 &  0.1 &     -112.413 &     -115.687 &      36.708 &      78.979 &   0.055015 \\
      CNN.T &  INFO-VAE &         - &      - &    10 &  5.0 &     -111.122 &     -122.567 &       8.437 &     114.130 &   0.004309 \\
      CNN.T &  INFO-VAE &         - &      - &    10 &  0.5 &     -103.501 &     -106.251 &      24.296 &      81.955 &   0.042997 \\
      CNN.T &  INFO-VAE &         - &      - &     1 &  0.1 &     -102.699 &     -105.651 &      37.987 &      67.665 &   0.109114 \\
      CNN.T &  INFO-VAE &         - &      - &    10 &  2.0 &     -101.911 &     -107.297 &      14.748 &      92.550 &   0.009465 \\
      CNN.T &    FB-VAE &        40 &      - &     - &      - &     -101.823 &     -104.936 &      37.671 &      67.265 &   0.131946 \\
      CNN.T &  INFO-VAE &         - &      - &    10 &  1.0 &     -101.392 &     -103.785 &      19.484 &      84.300 &   0.016348 \\
      CNN.T &  INFO-VAE &         - &      - &   100 &  0.1 &      -99.630 &     -102.594 &      34.361 &      68.233 &   0.003865 \\
      CNN.T &  BETA-VAE &         - &  5.00 &     - &      - &      -99.506 &     -111.420 &       9.002 &     102.418 &   0.001213 \\
      CNN.T &  INFO-VAE &         - &      - &   100 &  5.0 &      -99.419 &     -111.431 &       9.074 &     102.356 &   0.001087 \\
      CNN.T &  INFO-VAE &         - &      - &     1 &  5.0 &      -96.905 &     -111.545 &       8.959 &     102.586 &   0.001620 \\
      CNN.T &    FB-VAE &        32 &      - &     - &      - &      -96.250 &      -99.154 &      30.607 &      68.547 &   0.054848 \\
      CNN.T &  BETA-VAE &         - &  0.25 &     - &      - &      -95.583 &      -98.752 &      30.788 &      67.964 &   0.054161 \\
 PixelCNN++ &  INFO-VAE &         - &      - &    10 &  0.1 &      -92.755 &      -96.713 &      35.803 &      60.910 &   0.021386 \\
 PixelCNN++ &  INFO-VAE &         - &      - &     1 &  0.1 &      -92.494 &      -96.655 &      46.089 &      50.566 &   0.063386 \\
      CNN.T &  BETA-VAE &         - &  2.00 &     - &      - &      -92.004 &      -95.649 &      15.915 &      79.735 &   0.006296 \\
      CNN.T &  BETA-VAE &         - &  0.50 &     - &      - &      -91.366 &      -95.277 &      25.682 &      69.595 &   0.021536 \\
      CNN.T &  INFO-VAE &         - &      - &     1 &  1.0 &      -91.075 &      -93.643 &      20.741 &      72.901 &   0.010445 \\
      CNN.T &  INFO-VAE &         - &      - &   100 &  0.5 &      -91.047 &      -94.688 &      25.119 &      69.568 &   0.002931 \\
      CNN.T &  INFO-VAE &         - &      - &   100 &  1.0 &      -90.987 &      -93.405 &      20.492 &      72.913 &   0.002752 \\
      CNN.T &  INFO-VAE &         - &      - &     1 &  2.0 &      -90.903 &      -95.517 &      15.738 &      79.779 &   0.004205 \\
      CNN.T &  BETA-VAE &         - &  0.75 &     - &      - &      -90.609 &      -93.535 &      22.650 &      70.885 &   0.014209 \\
      CNN.T &    FB-VAE &        24 &      - &     - &      - &      -90.534 &      -94.576 &      24.228 &      70.348 &   0.019529 \\
      CNN.T &    FB-VAE &         4 &      - &     - &      - &      -90.221 &      -93.075 &      20.691 &      72.385 &   0.009502 \\
      CNN.T &    FB-VAE &         8 &      - &     - &      - &      -90.140 &      -93.111 &      20.838 &      72.272 &   0.012566 \\
      CNN.T &    FB-VAE &        16 &      - &     - &      - &      -89.955 &      -93.511 &      21.312 &      72.199 &   0.012732 \\
      CNN.T &  INFO-VAE &         - &      - &   100 &  2.0 &      -89.805 &      -95.291 &      15.892 &      79.399 &   0.002115 \\
      CNN.T &  BETA-VAE &         - &  1.00 &     - &      - &      -89.767 &      -93.170 &      20.516 &      72.654 &   0.009939 \\
      CNN.T &  INFO-VAE &         - &      - &     1 &  0.5 &      -89.740 &      -94.728 &      25.616 &      69.112 &   0.020474 \\
 PixelCNN++ &    FB-VAE &        40 &      - &     - &      - &      -89.652 &      -91.937 &      38.192 &      53.746 &   0.088447 \\
      CNN.T &  BETA-VAE &         - &  1.50 &     - &      - &      -88.192 &      -93.665 &      18.052 &      75.612 &   0.006233 \\
 PixelCNN++ &  BETA-VAE &         - &  0.25 &     - &      - &      -88.009 &      -90.369 &      38.595 &      51.773 &   0.026900 \\
 PixelCNN++ &    FB-VAE &        32 &      - &     - &      - &      -83.701 &      -86.769 &      31.107 &      55.661 &   0.024090 \\
 PixelCNN++ &    FB-VAE &        24 &      - &     - &      - &      -83.304 &      -84.196 &      23.840 &      60.356 &   0.008685 \\
 PixelCNN++ &  INFO-VAE &         - &      - &    10 &  5.0 &      -83.249 &      -82.019 &       0.000 &      82.018 &  -0.000051 \\
 PixelCNN++ &  INFO-VAE &         - &      - &    10 & 10.0 &      -83.124 &      -82.152 &       0.000 &      82.152 &   0.000044 \\
 PixelCNN++ &  INFO-VAE &         - &      - &    10 &  2.0 &      -82.733 &      -81.971 &       0.000 &      81.970 &  -0.000009 \\
 PixelCNN++ &  INFO-VAE &         - &      - &    10 &  0.5 &      -82.244 &      -82.234 &       0.002 &      82.232 &   0.000076 \\
 PixelCNN++ &  INFO-VAE &         - &      - &    10 &  1.0 &      -82.243 &      -82.169 &       0.001 &      82.169 &  -0.000157 \\
 PixelCNN++ &  INFO-VAE &         - &      - &     1 &  5.0 &      -81.965 &      -80.716 &       0.000 &      80.716 &   0.000060 \\
 PixelCNN++ &  BETA-VAE &         - &  0.50 &     - &      - &      -81.886 &      -84.875 &      26.952 &      57.923 &   0.006013 \\
 PixelCNN++ &  INFO-VAE &         - &      - &     1 & 10.0 &      -81.783 &      -82.144 &       0.000 &      82.144 &  -0.000015 \\
 PixelCNN++ &  BETA-VAE &         - &  1.50 &     - &      - &      -81.690 &      -80.745 &       0.000 &      80.745 &  -0.000073 \\
 PixelCNN++ &  INFO-VAE &         - &      - &   100 &  1.0 &      -81.587 &      -80.926 &       0.003 &      80.923 &  -0.000021 \\
 PixelCNN++ &  INFO-VAE &         - &      - &   100 &  0.1 &      -81.552 &      -80.771 &       0.004 &      80.767 &   0.000193 \\
 PixelCNN++ &  INFO-VAE &         - &      - &   100 &  5.0 &      -81.419 &      -80.846 &       0.003 &      80.843 &   0.000667 \\
 PixelCNN++ &  BETA-VAE &         - &  1.00 &     - &      - &      -81.165 &      -80.664 &       0.001 &      80.663 &   0.000035 \\
 PixelCNN++ &  INFO-VAE &         - &      - &   100 & 10.0 &      -81.058 &      -81.975 &       0.001 &      81.974 &  -0.000118 \\
 PixelCNN++ &  INFO-VAE &         - &      - &   100 &  2.0 &      -80.952 &      -80.756 &       0.011 &      80.744 &   0.000926 \\
 PixelCNN++ &    FB-VAE &        16 &      - &     - &      - &      -80.698 &      -81.895 &      16.927 &      64.968 &   0.002087 \\
 PixelCNN++ &  INFO-VAE &         - &      - &     1 &  2.0 &      -80.670 &      -80.555 &       0.003 &      80.553 &   0.000333 \\
 PixelCNN++ &  INFO-VAE &         - &      - &     1 &  0.5 &      -80.669 &      -84.514 &      26.190 &      58.324 &   0.003956 \\
 PixelCNN++ &  INFO-VAE &         - &      - &   100 &  0.5 &      -80.658 &      -80.769 &       0.003 &      80.766 &   0.000000 \\
 PixelCNN++ &  BETA-VAE &         - &  2.00 &     - &      - &      -80.505 &      -80.817 &       0.000 &      80.817 &   0.000025 \\
 PixelCNN++ &    FB-VAE &         4 &      - &     - &      - &      -80.446 &      -80.548 &       5.755 &      74.794 &   0.000707 \\
 PixelCNN++ &  BETA-VAE &         - &  5.00 &     - &      - &      -80.228 &      -80.670 &       0.000 &      80.669 &   0.000144 \\
 PixelCNN++ &  INFO-VAE &         - &      - &     1 &  1.0 &      -80.023 &      -80.621 &       0.000 &      80.620 &   0.000174 \\
 PixelCNN++ &  BETA-VAE &         - &  0.75 &     - &      - &      -79.370 &      -80.819 &       8.785 &      72.034 &   0.000669 \\
 PixelCNN++ &  BETA-VAE &         - & 10.00 &     - &      - &      -79.335 &      -80.604 &       0.000 &      80.604 &  -0.000003 \\
 PixelCNN++ &    FB-VAE &         8 &      - &     - &      - &      -78.865 &      -80.663 &       9.761 &      70.902 &   0.001080 \\
\bottomrule
    \end{tabular}
    \caption{Full intrinsic evaluation results of experiments on Binarised MNIST dataset}
    \label{tab:bmnist_full_results}
\end{table*}

\begin{table*}[!htb]
    \centering
    \scriptsize
    \begin{tabular}{llllll|rrrrr}
    \toprule
        Decoder & Objective & $\lambda_{\text{FB}}$ &  $\beta$ & $\lambda_{\text{MMD}}$ & $\lambda_{\text{rate}}$ &    IW LL &     ELBO &    Rate &  Distortion &        MMD \\
\midrule
 Distil roBERTa &  BETA-VAE &         - & 0.000 &     - &      - & -253.314 & -262.025 & 199.762 &      62.263 &   0.004164 \\
 Distil roBERTa &    FB-VAE &        64 &     - &     - &      - & -120.653 & -125.793 &  60.116 &      65.677 &   0.000889 \\
 Distil roBERTa &  INFO-VAE &         - &     - &    10 &  0.1 & -115.817 & -119.895 &  51.583 &      68.311 &   0.000932 \\
 Distil roBERTa &  INFO-VAE &         - &     - &   100 &  0.1 & -114.690 & -118.358 &  48.643 &      69.715 &   0.000672 \\
 Distil roBERTa &  INFO-VAE &         - &     - &     1 &  0.1 & -111.780 & -119.874 &  51.270 &      68.604 &   0.000934 \\
 Distil roBERTa &    FB-VAE &        32 &     - &     - &      - &  -97.405 & -103.309 &  30.311 &      72.998 &   0.000347 \\
 Distil roBERTa &  BETA-VAE &         - & 0.25 &     - &      - &  -96.893 & -101.637 &  26.318 &      75.319 &   0.000433 \\
 Distil roBERTa &    FB-VAE &        16 &     - &     - &      - &  -94.325 &  -95.389 &  15.632 &      79.757 &   0.000246 \\
 Distil roBERTa &  BETA-VAE &         - & 1.00 &     - &      - &  -93.939 &  -89.989 &   0.006 &      89.983 &   0.000011 \\
 Distil roBERTa &  INFO-VAE &         - &     - &    10 &  2.0 &  -92.038 &  -90.203 &   0.002 &      90.202 &  -0.000003 \\
 Distil roBERTa &  BETA-VAE &         - & 2.00 &     - &      - &  -91.553 &  -90.596 &   0.001 &      90.595 &   0.000004 \\
 Distil roBERTa &  INFO-VAE &         - &     - &     1 &  2.0 &  -90.949 &  -90.400 &   0.001 &      90.399 &   0.000003 \\
 Distil roBERTa &    FB-VAE &         8 &     - &     - &      - &  -90.527 &  -92.318 &   8.047 &      84.271 &   0.000134 \\
 Distil roBERTa &  INFO-VAE &         - &     - &    10 &  0.5 &  -90.219 &  -91.485 &   4.573 &      86.912 &   0.000085 \\
 Distil roBERTa &  INFO-VAE &         - &     - &     1 &  1.0 &  -90.100 &  -90.172 &   0.004 &      90.168 &   0.000018 \\
 Distil roBERTa &  INFO-VAE &         - &     - &    10 &  1.0 &  -90.073 &  -90.215 &   0.005 &      90.210 &  -0.000010 \\
 Distil roBERTa &    FB-VAE &         4 &     - &     - &      - &  -89.834 &  -91.519 &   3.829 &      87.690 &   0.000075 \\
 Distil roBERTa &  BETA-VAE &         - & 0.75 &     - &      - &  -89.629 &  -90.604 &   0.021 &      90.583 &   0.000017 \\
 Distil roBERTa &  BETA-VAE &         - & 0.50 &     - &      - &  -89.176 &  -91.301 &   4.669 &      86.632 &   0.000131 \\
 Distil roBERTa &  INFO-VAE &         - &     - &     1 &  0.5 &  -88.973 &  -91.410 &   4.468 &      86.943 &   0.000075 \\
 Distil roBERTa &  INFO-VAE &         - &     - &   100 &  2.0 &  -88.214 &  -90.436 &   0.002 &      90.434 &   0.000004 \\
 Distil roBERTa &  INFO-VAE &         - &     - &   100 &  0.5 &  -87.757 &  -91.146 &   3.669 &      87.477 &   0.000156 \\
 Distil roBERTa &  INFO-VAE &         - &     - &   100 &  1.0 &  -86.616 &  -90.574 &   0.004 &      90.570 &   0.000018 \\
\bottomrule
    \end{tabular}
    \caption{Full intrinsic evaluation results of experiments on Penn Treebank dataset}
    \label{tab:ptb_full_results}
\end{table*}

\section{Grouped mixed membership models}\label{app:dp-models}

To compare the statistic histograms across groups in a systematic way, we fit Dirichlet process mixed membership models to the scalar valued statistics with the group variable denoting the different experiments and control group. In total we fit 9 of these models for the three statistics across three latent structure models. We use a truncated Normal as family for the components and uniform priors for the location and scale parameters of these components. The number of components is 5 for both the statistics resulting from the sequence length model and topic model and 7 for the MNIST digit identity model.

\begin{figure*}[!htb]
    \centering
    \includegraphics[width=\textwidth]{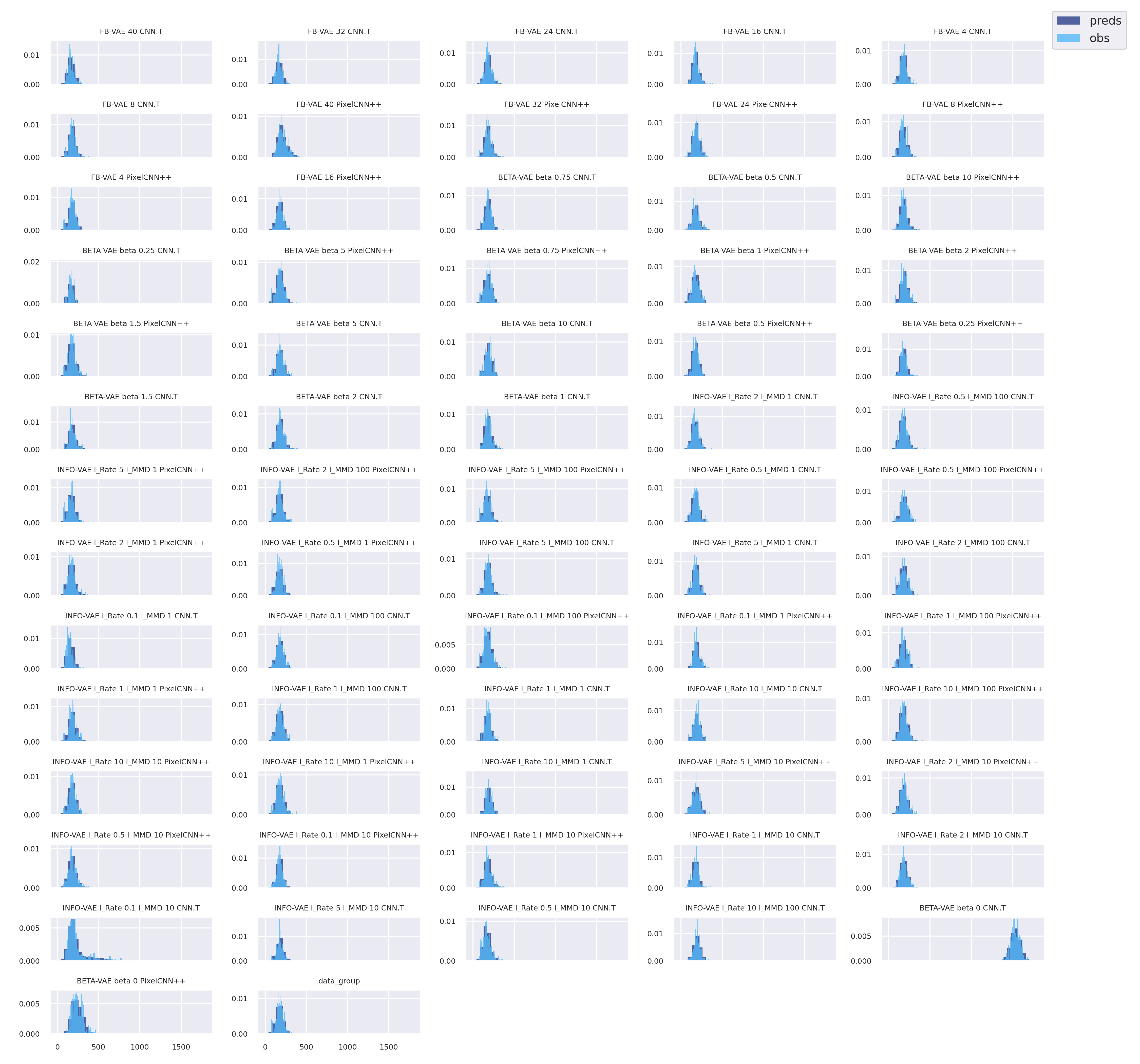}
    \caption{Posterior predictive samples (preds) versus observations (obs) of the $T(x^*)$ statistic assessed under the MNIST digit identity latent structure model as modelled by the DP mixture model.}
    \label{fig:surprisal_check_mnist_uncon_uncon}
\end{figure*}

\begin{figure*}[!htb]
    \centering
    \includegraphics[width=\textwidth]{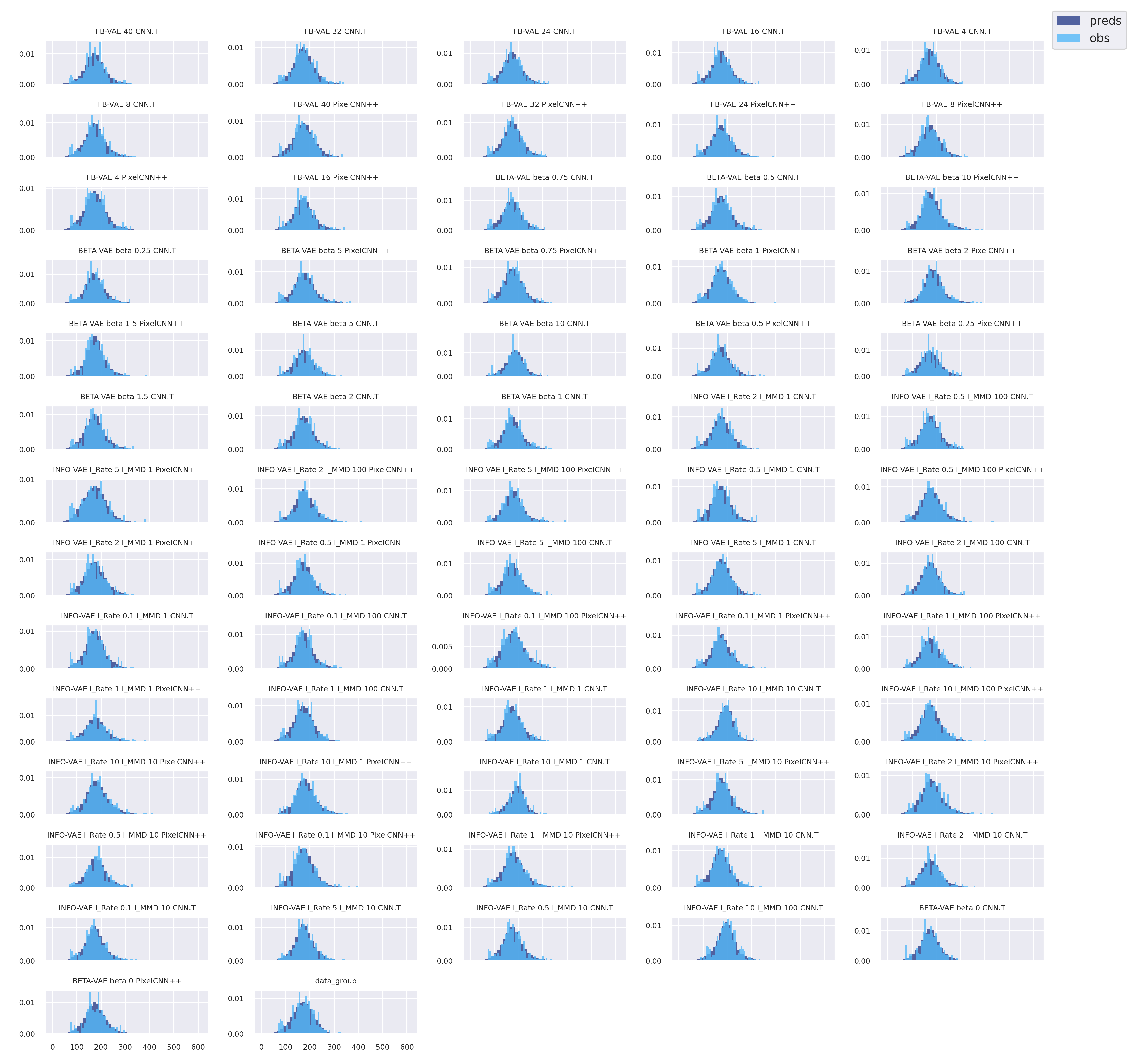}
    \caption{Posterior predictive samples (preds) versus observations (obs) of the $T(\tilde x^*)$ statistic assessed under the MNIST digit identity latent structure model as modelled by the DP mixture model.}
    \label{fig:surprisal_check_mnist_uncon_con}
\end{figure*}

\begin{figure*}[!htb]
    \centering
    \includegraphics[width=\textwidth]{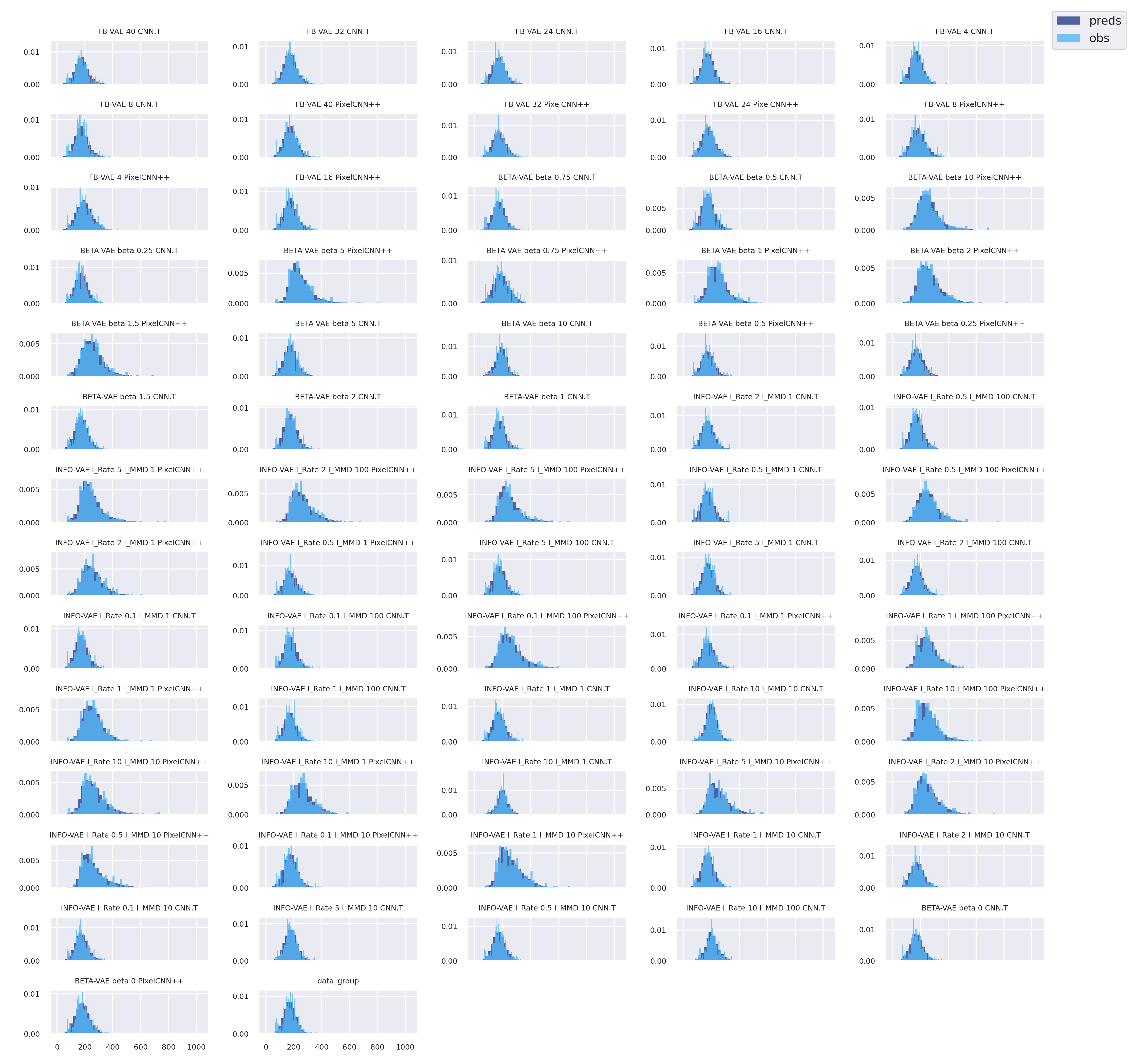}
    \caption{Posterior predictive samples (preds) versus observations (obs) of the $T(\tilde x^* | x^*)$ statistic assessed under the MNIST digit identity latent structure model as modelled by the DP mixture model.}
    \label{fig:surprisal_check_mnist_con_con}
\end{figure*}

\begin{figure*}[!htb]
    \centering
    \includegraphics[width=\textwidth]{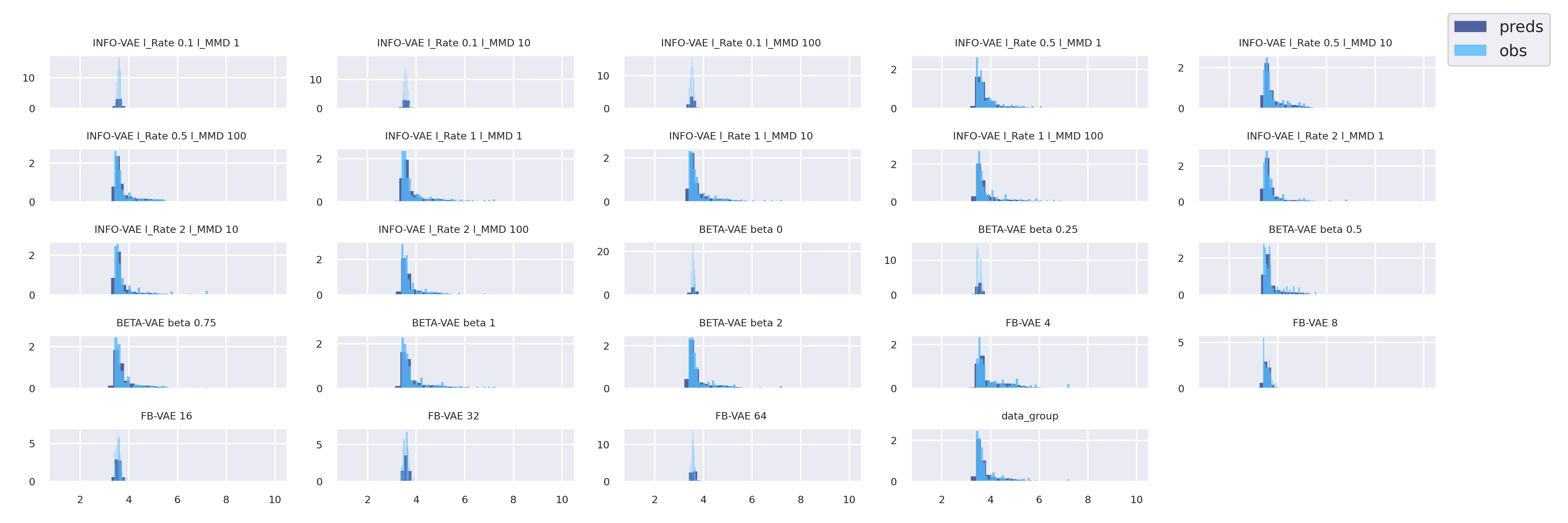}
    \caption{Posterior predictive samples (preds) versus observations (obs) of the $T(x^*)$ statistic assessed under the Penn Treebank sequence length latent structure model as modelled by the DP mixture model.}
    \label{fig:surprisal_check_ptb_seq_len_uncon_uncon}
\end{figure*}

\begin{figure*}[!htb]
    \centering
    \includegraphics[width=\textwidth]{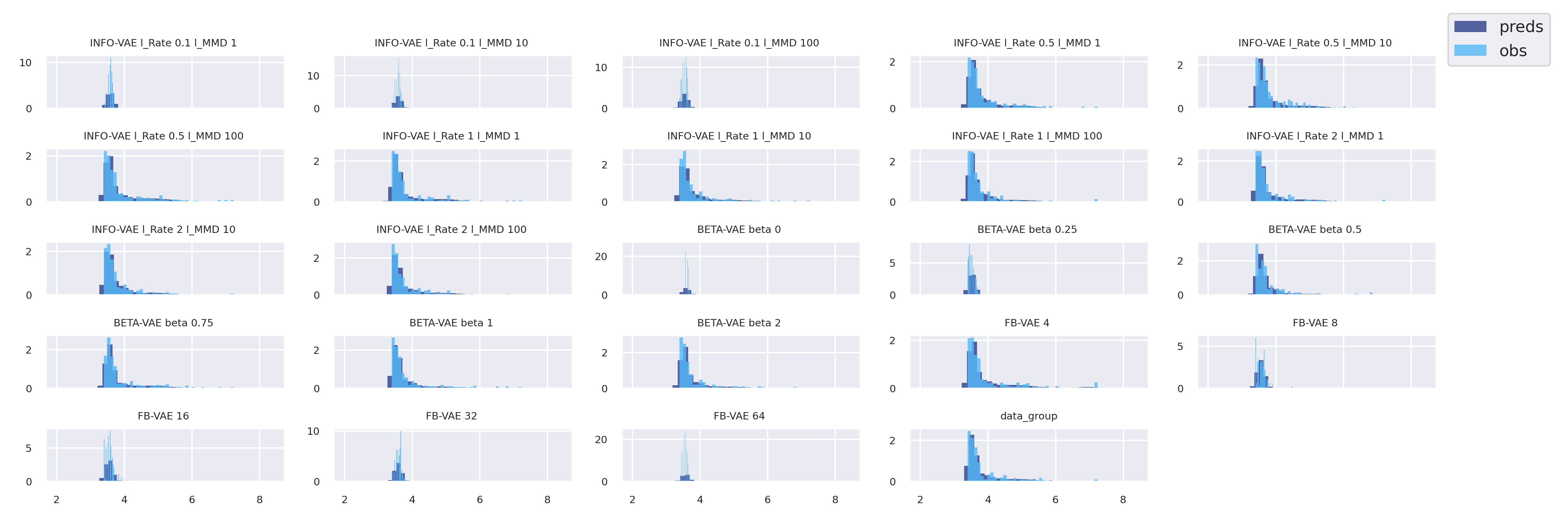}
    \caption{Posterior predictive samples (preds) versus observations (obs) of the $T(\tilde x^*)$ statistic assessed under the Penn Treebank sequence length latent structure model as modelled by the DP mixture model.}
    \label{fig:surprisal_check_ptb_seq_len_uncon_con}
\end{figure*}

\begin{figure*}[!htb]
    \centering
    \includegraphics[width=\textwidth]{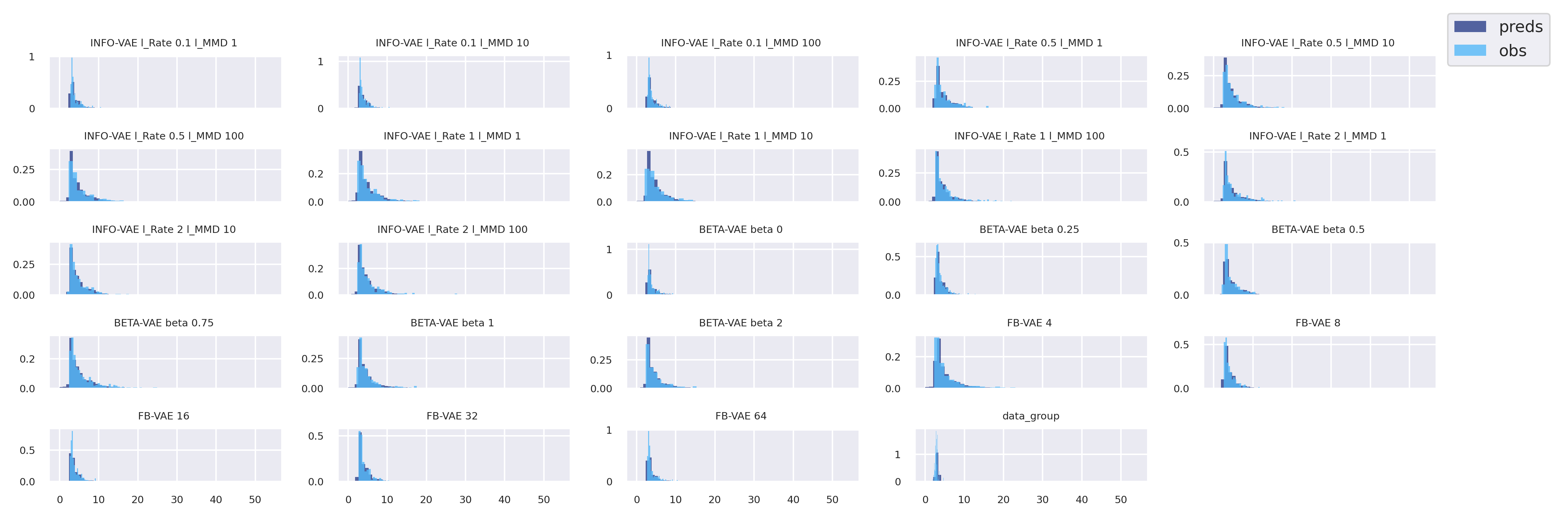}
    \caption{Posterior predictive samples (preds) versus observations (obs) of the $T(\tilde x^*|x^*)$ statistic assessed under the Penn Treebank sequence length latent structure model as modelled by the DP mixture model.}
    \label{fig:surprisal_check_ptb_seq_len_con_con}
\end{figure*}

\begin{figure*}[!htb]
    \centering
    \includegraphics[width=\textwidth]{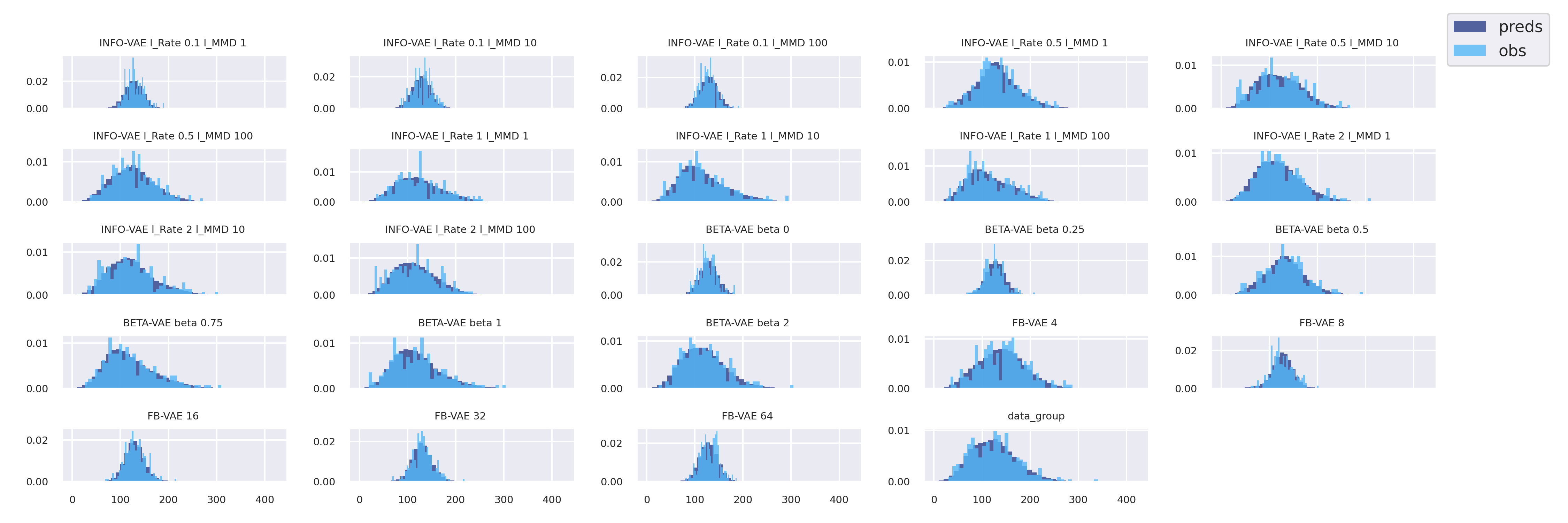}
    \caption{Posterior predictive samples (preds) versus observations (obs) of the $T(x^*)$ statistic assessed under the Penn Treebank topic latent structure model as modelled by the DP mixture model.}
    \label{fig:surprisal_check_ptb_topics_uncon_uncon}
\end{figure*}

\begin{figure*}[!htb]
    \centering
    \includegraphics[width=\textwidth]{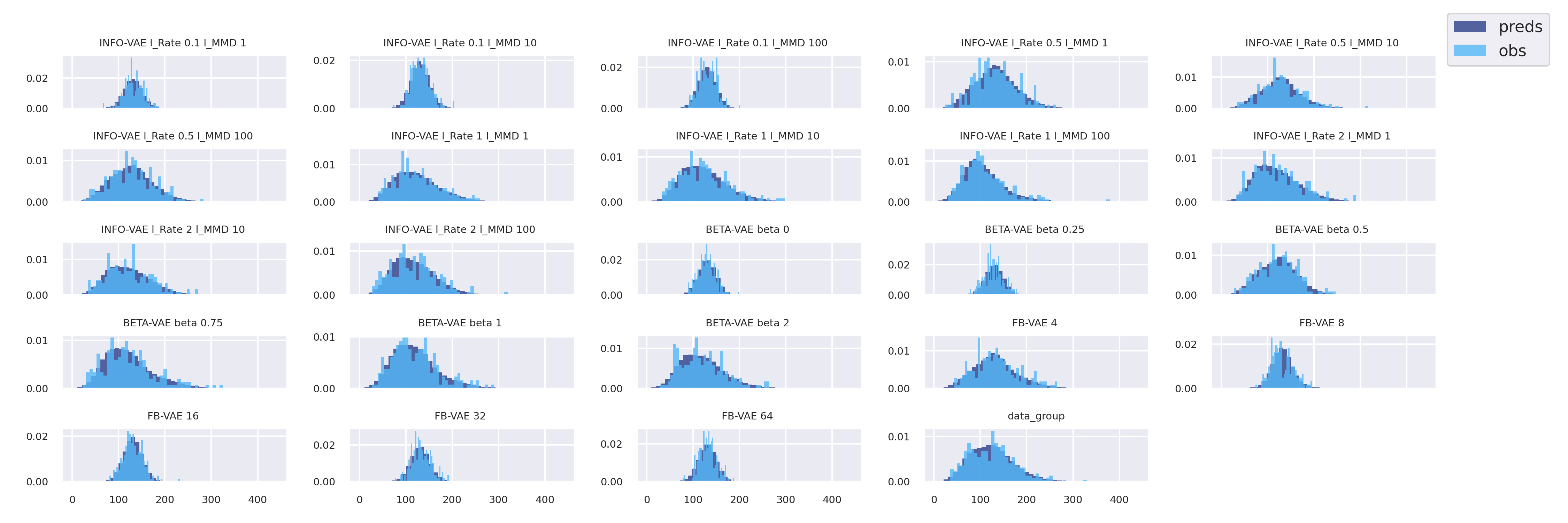}
    \caption{Posterior predictive samples (preds) versus observations (obs) of the $T(\tilde x^*)$ statistic assessed under the Penn Treebank topic latent structure model as modelled by the DP mixture model.}
    \label{fig:surprisal_check_ptb_topics_uncon_con}
\end{figure*}

\begin{figure*}[!htb]
    \centering
    \includegraphics[width=\textwidth]{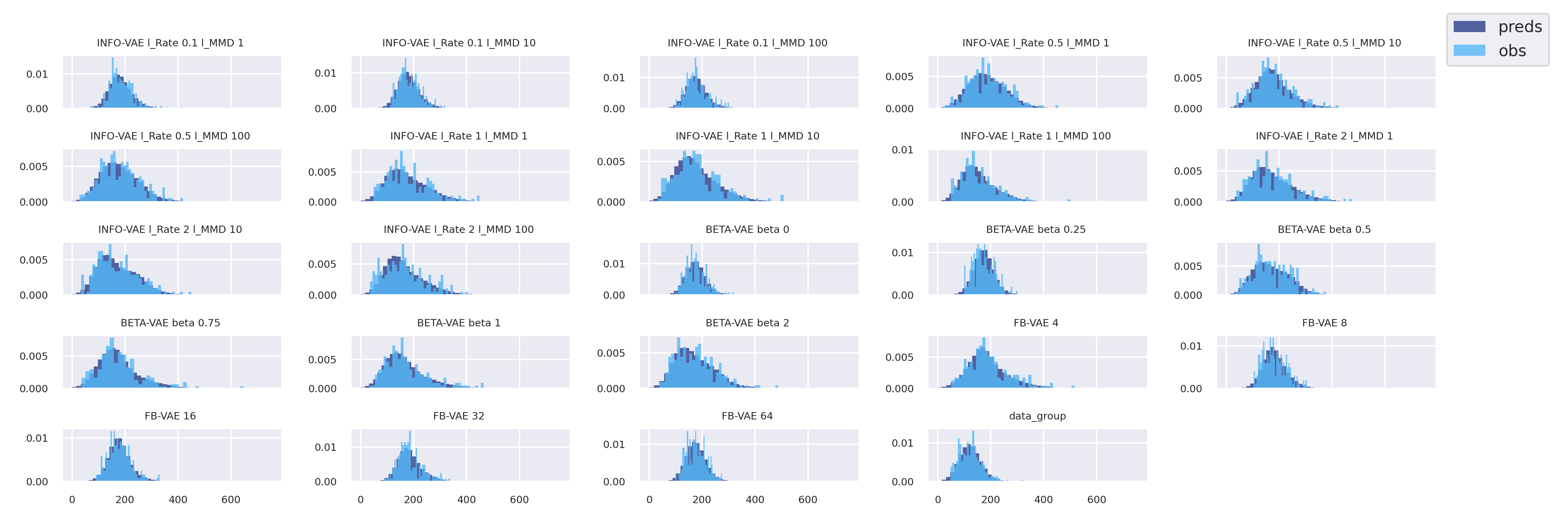}
    \caption{Posterior predictive samples (preds) versus observations (obs) of the $T(\tilde x^*|x^*)$ statistic assessed under the Penn Treebank topic latent structure model as modelled by the DP mixture model.}
    \label{fig:surprisal_check_ptb_topics_con_con}
\end{figure*}

Figures \ref{fig:surprisal_check_mnist_uncon_uncon} -- \ref{fig:surprisal_check_ptb_topics_con_con} depict posterior predictive samples for different groups under the analysis DP mixture models that are used to compare the statistics  $T(x^*)$,  $T(\tilde x^*)$ and $T(\tilde x^*|x^*)$ evaluated under the three latent structure models (the MNIST digit identity model, the Penn Treebank sequence length model and the Penn Treebank LDA topic model) for different groups. 

\section{Estimated divergence to control group}\label{app:kl-plots}

\begin{figure*}[!htb]
    \centering
    \includegraphics[width=0.8\textwidth]{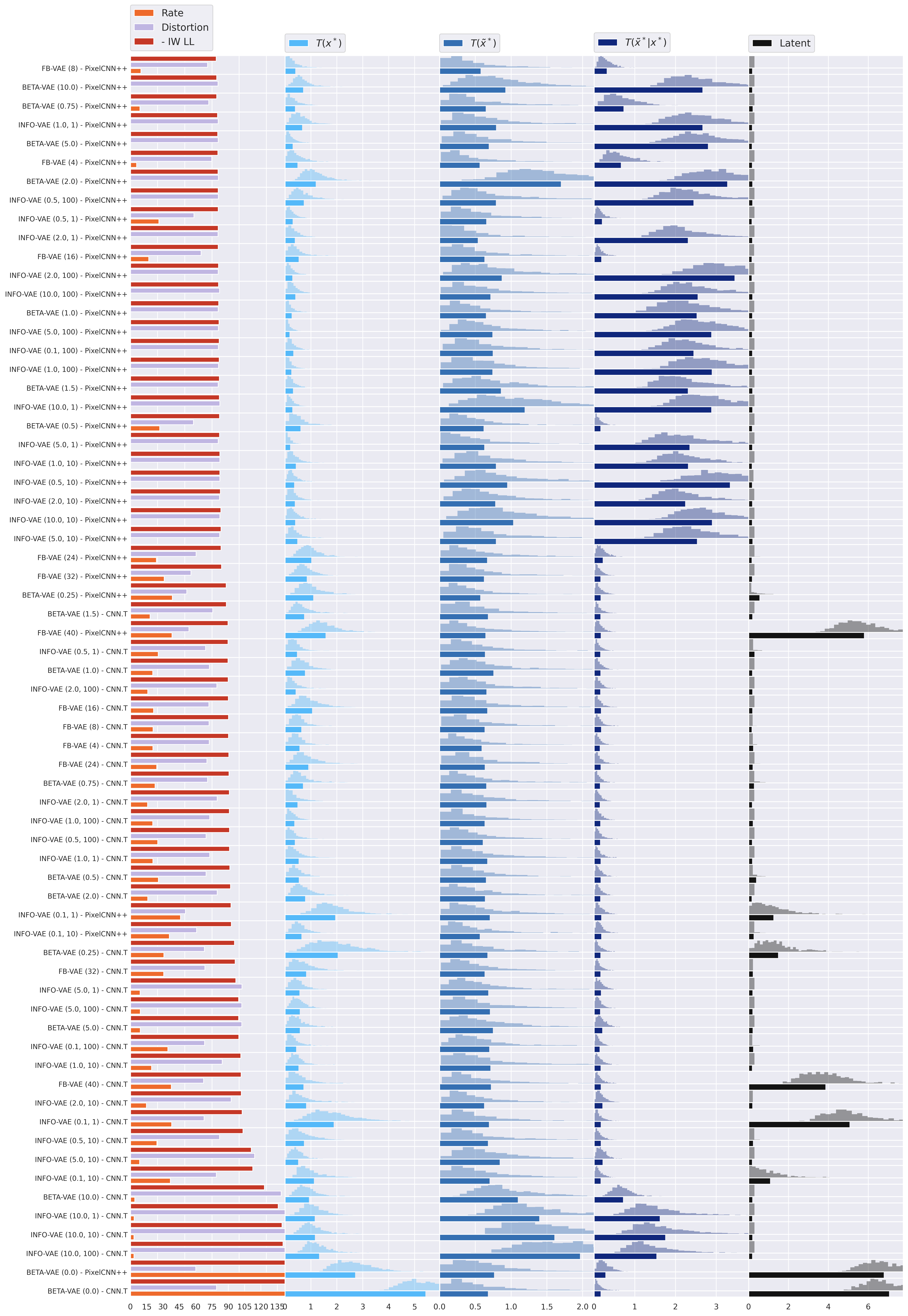}
    \caption{Full experimental results for the control group divergence analysis for the MNIST digit identity structure model. The left most column shows the intrinsic evaluation metrics for reference. The middle three columns show estimated divergences from the control group under our analysis model. The right most column shows the control group divergence under the latent analysis model. The horizontal bars denote the average value of the sampled divergences plotted as histograms. The experiments are labelled with the objectives according to the following format: \infovae ($\lambda_{\text{rate}}$, $\lambda_{\text{MMD}}$), \betavae ($\beta$) and \fbvae ($\lambda_{\text{FB}}$). We additionally distinguish between decoder types used: CNN.T or PixelCNN++.}
    \label{fig:kl-plot-mnist}
\end{figure*}

\begin{figure*}[!htb]
    \centering
    \includegraphics[width=\textwidth]{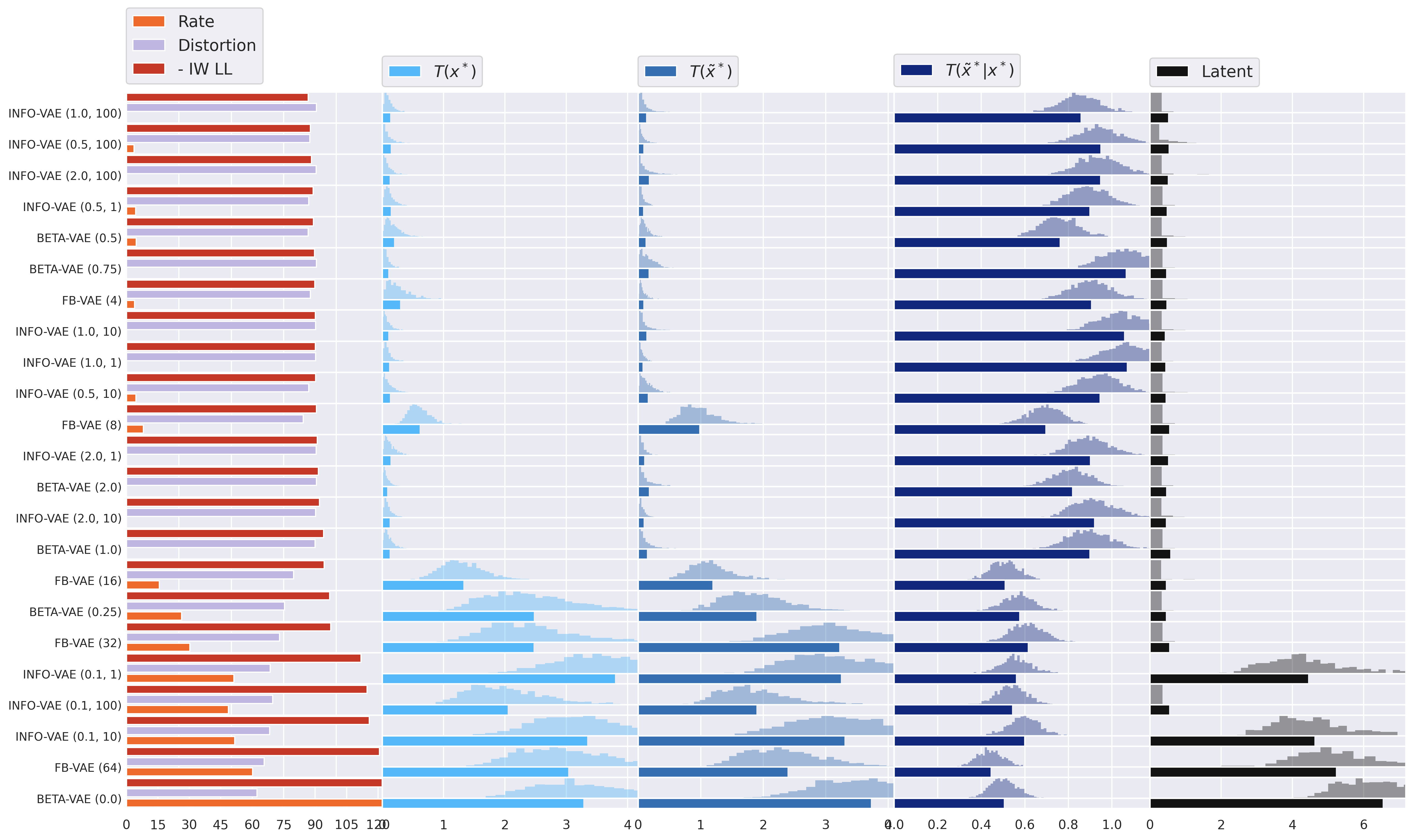}
    \caption{Full experimental results for the control group divergence analysis for the PTB sequence length latent structure model. The left most column shows the intrinsic evaluation metrics for reference. The middle three columns show estimated divergences from the control group under our analysis model. The right most column shows the control group divergence under the latent analysis model. The horizontal bars denote the average value of the sampled divergences plotted as histograms. The experiments are labelled with the objectives according to the following format: \infovae ($\lambda_{\text{rate}}$, $\lambda_{\text{MMD}}$), \betavae ($\beta$) and \fbvae ($\lambda_{\text{FB}}$). }
    \label{fig:kl-plot-ptb-seq-len}
\end{figure*}

\begin{figure*}[!htb]
    \centering
    \includegraphics[width=\textwidth]{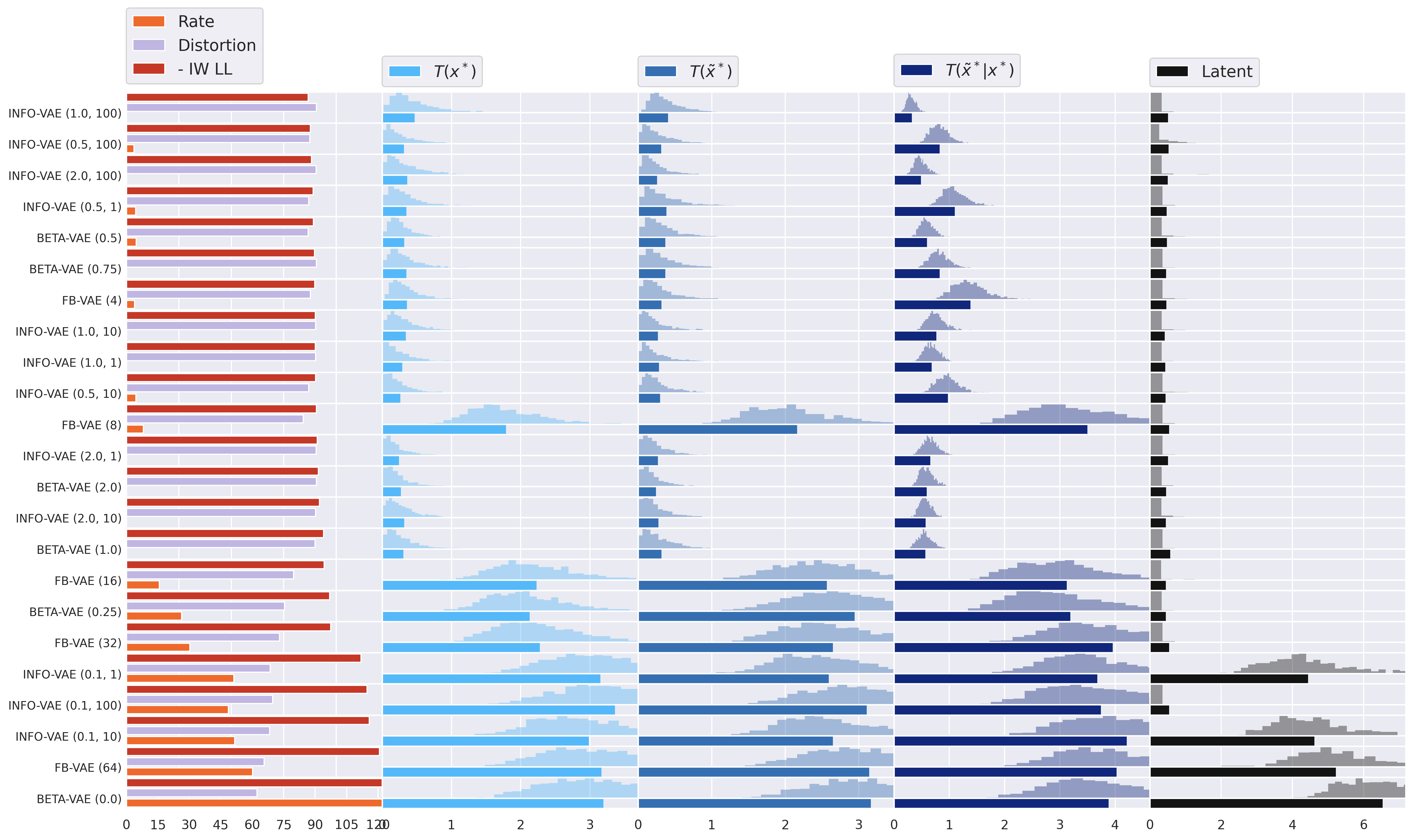}
    \caption{Full experimental results for the control group divergence analysis for the PTB topic structure model. The left most column shows the intrinsic evaluation metrics for reference. The middle three columns show estimated divergences from the control group under our analysis model. The right most column shows the control group divergence under the latent analysis model. The horizontal bars denote the average value of the sampled divergences plotted as histograms. The experiments are labelled with the objectives according to the following format: \infovae ($\lambda_{\text{rate}}$, $\lambda_{\text{MMD}}$), \betavae ($\beta$) and \fbvae ($\lambda_{\text{FB}}$). }
    \label{fig:kl-plot-ptb-topics}
\end{figure*}

In Figures \ref{fig:kl-plot-mnist} -- \ref{fig:kl-plot-ptb-topics} we plot full experimental results of the plots equivalent to those presented in Figure \ref{fig:all-kl-plots} in Section \ref{sec:experiments}.

\end{document}